%% file: sample_paper.tex
\documentclass[twoside]{article}

\usepackage[accepted]{aistats2021}
\usepackage[utf8]{inputenc} %
\usepackage[T1]{fontenc}    %
\usepackage{url}            %
\usepackage{booktabs}       %
\usepackage{amsfonts}       %
\usepackage{nicefrac}       %
\usepackage{microtype}      %

\usepackage{natbib}
\usepackage{graphicx}
\usepackage{booktabs} %

\input{math_commands}

\usepackage{slashbox}
\usepackage{framed}
\usepackage{algorithm}
\usepackage{algorithmicx}
\usepackage{algpseudocode}
\usepackage{wrapfig}
\usepackage{lipsum}
\usepackage{url}
\usepackage{caption}
\usepackage{graphicx}
\usepackage{subcaption}
\usepackage{comment}
\usepackage{graphicx}
\usepackage{xcolor}
\usepackage[font=small,labelfont=bf]{caption}
\usepackage{booktabs}       %
\usepackage{tabularx}
\usepackage{mathtools}
\usepackage{multirow}
\usepackage{xcolor}
\usepackage[colorlinks=true]{hyperref}
\usepackage{bm}

\usepackage{amsthm}
\newtheorem{theorem}{Theorem}
\newtheorem{remark}{Remark}
\usepackage{chngcntr}
\counterwithin*{theorem}{subsection}

\definecolor{mydarkblue}{rgb}{0,0.08,0.45}
\definecolor{darkgreen}{RGB}{34,139,34}

\definecolor{myblue}{RGB}{49,130,189}
\definecolor{myred}{RGB}{251,106,74}

\hypersetup{colorlinks=true,
    linkcolor=mydarkblue,
    citecolor=mydarkblue,
    filecolor=mydarkblue,
    urlcolor=mydarkblue}

\usepackage{scalerel,stackengine}
\stackMath
\newcommand\reallywidehat[1]{%
\savestack{\tmpbox}{\stretchto{%
  \scaleto{%
    \scalerel*[\widthof{\ensuremath{#1}}]{\kern.1pt\mathchar"0362\kern.1pt}%
    {\rule{0ex}{\textheight}}%
  }{\textheight}%
}{2.4ex}}%
\stackon[-6.9pt]{#1}{\tmpbox}%
}

\usepackage{tikz}
\newcommand*\circled[1]{\tikz[baseline=(char.base)]{
            \node[shape=circle,draw,inner sep=2pt] (char) {#1};}}

\begin{document}

\twocolumn[
\aistatstitle{Beyond Marginal Uncertainty: How Accurately can Bayesian \\ Regression Models Estimate Posterior Predictive Correlations?}

\aistatsauthor{ Chaoqi Wang$^\star$ \\ University of Chicago \And Shengyang Sun$^\star$  \\ University of Toronto\\Vector Institute \And  Roger Grosse \\ University of Toronto\\Vector Institute }
\aistatsaddress{}]

\begin{abstract}

While uncertainty estimation is a well-studied topic in deep learning, most such work focuses on \emph{marginal} uncertainty estimates, i.e.~the predictive mean and variance at individual input locations. But it is often more useful to estimate predictive \emph{correlations} between the function values at \emph{different} input locations. In this paper, we consider the problem of benchmarking how accurately Bayesian models can estimate predictive correlations. We first consider a downstream task which depends on posterior predictive correlations: transductive active learning (TAL). We find that TAL makes better use of models' uncertainty estimates than ordinary active learning, and recommend this as a benchmark for evaluating Bayesian models. Since TAL is too expensive and indirect to guide development of algorithms, we introduce two metrics which more directly evaluate the predictive correlations and which can be computed efficiently: meta-correlations (i.e.~the correlations between the models correlation estimates and the true values), and cross-normalized likelihoods (XLL). We validate these metrics by demonstrating their consistency with TAL performance and obtain insights about the relative performance of current Bayesian neural net and Gaussian process models. %
\end{abstract}

 \input{tex/introduction.tex}

\input{tex/preliminary.tex}

\input{tex/method.tex}

 \input{tex/related_works.tex}

\input{tex/active_learning.tex}

\input{tex/discussion_and_conclusion}

\section*{Acknowledgements}
We thank Saminul Haque, Guodong Zhang, Wenyuan Zeng and Cong Han Lim for their insightful comments and discussions on this project. We also thank the Vector Institute for providing the scientific computing resources. This research project has been supported by LG Electronics. SS was supported by the Connaught Fellowship. RG acknowledges support from the CIFAR Canadian AI Chairs program.

\bibliographystyle{citation}
\bibliography{bib/gp,bib/bnn,bib/gan,bib/vi,bib/misc,bib/bayesian,bib/nn,bib/generalization,bib/activeL,bib/bo,bib/rl,bib/kfac}

\appendix
\newpage
\input{tex/appendix.tex}

\end{document}

%% file: math_commands.tex
\usepackage{amsmath,amsfonts,bm}

\def\eqref#1{equation~\ref{#1}}

\def\1{\bm{1}}

\def\ry{{\textnormal{y}}}

\def\vzero{{\bm{0}}}

\def\vmu{{\bm{\mu}}}

\def\vd{{\bm{d}}}

\def\vf{{\bm{f}}}

\def\vr{{\bm{r}}}

\def\vw{{\bm{w}}}
\def\vx{{\bm{x}}}
\def\vy{{\bm{y}}}

\def\mA{{\bm{A}}}
\def\mB{{\bm{B}}}
\def\mC{{\bm{C}}}
\def\mD{{\bm{D}}}

\def\mF{{\bm{F}}}

\def\mI{{\bm{I}}}

\def\mX{{\bm{X}}}

\def\mSigma{{\bm{\Sigma}}}

\DeclareMathAlphabet{\mathsfit}{\encodingdefault}{\sfdefault}{m}{sl}
\SetMathAlphabet{\mathsfit}{bold}{\encodingdefault}{\sfdefault}{bx}{n}

\def\dataset{{\mathcal{D}}}

\newcommand{\KL}[2]{\mathrm{KL}\left(#1\|#2\right)}
\newcommand{\Var}{\mathrm{Var}}

\newcommand{\abs}[1]{\left|#1\right|}
\newcommand{\bone}{\mathbf{1}}
\newcommand{\normal}{\mathcal{N}}
\newcommand{\trace}{\mathrm{tr}}
\newcommand{\Reals}{\mathbb{R}}
\newcommand{\bx}{\mathbf{x}}
\newcommand{\bz}{\mathbf{z}}
\newcommand{\ba}{\mathbf{a}}

\newcommand{\bw}{\mathbf{w}}

\newcommand{\bC}{\mathbf{C}}
\newcommand{\bI}{\mathbb{I}}
\newcommand{\bmu}{\bm{\mu}}
\newcommand{\bSigma}{\bm{\Sigma}}
\newcommand{\bsigma}{\bm{\sigma}}

\newcommand{\mean}{{\boldsymbol \mu}}
\newcommand{\data}{\mathcal{D}}

\newcommand{\expect}{\mathbb{E}}

\DeclareMathOperator{\diag}{diag}

%% file: tex/introduction.tex
\section{Introduction}

Uncertainty estimation is a key desideratum for modern deep learning systems, and is essential for guiding exploration and making robust decisions. %
Most works on uncertainty estimation have focused on \emph{marginal uncertainty}, i.e.~the posterior predictive variance $\Var(f(x) | \mathcal{D}_{tr})$ of a function $f$ at a single location $x$ conditioned on the training data $\mathcal{D}_{tr}$. Marginal uncertainty is used in many active learning and Bayesian optimization algorithms~\citep{hernandez2015probabilistic, guo2017calibration, cobb2018loss}, as high uncertainty at a point indicates it is favorable to query.

However, some algorithms can explore even more efficiently by exploiting \emph{posterior predictive correlations} (PPCs) between function values at multiple locations. Mathematically, we are interested in $\rho(f(x), f(x') | \mathcal{D}_{tr})$, where $\rho$ denotes the Pearson correlation. Algorithms that exploit PPCs include transductive experiment design \citep{yu2006active}, where the goal is to acquire information relevant to a specific prediction $f(x_\star)$; in this case, it is advantageous to query locations whose values have a high PPC with $f(x_\star)$. Similarly, in cost-sensitive Bayesian optimization, it would be very useful to make cheap queries in order to indirectly learn about more expensive (and likely higher-performing) regions \citep{hennig2012entropy, frazier2009knowledge}. Arguably, any task requiring exploration directed towards a specific goal should benefit from accurately modeling dependencies between function values that are (cheaply) observable and also relevant to the goal.  %

Despite its importance, the problem of estimating PPCs has received little attention in machine learning, and applications of PPCs currently rely on a handful of models for which they can be tractably computed, such as Gaussian processes. We believe this neglect is due to PPC estimates being much harder to evaluate than marginal uncertainty estimates. For evaluating marginal uncertainty, predictive log-likelihoods and reliability plots \citep{guo2017calibration} are widely accepted tools. However, no analogous metrics exist for evaluating PPCs. In this paper, we introduce and validate three metrics for evaluating PPCs under different considerations and use them to benchmark various Bayesian models.

Conceptually, the most straightforward way to evaluate PPC estimators is by comparing their estimates to the true PPC values. We formalize this by measuring \emph{metacorrelations}, i.e.~the correlations between a model's PPC estimates and the true values. Unfortunately, metacorrelations have limited applicability because the true PPC values are only defined for synthetic data drawn from a known distribution. However, they are very useful as a tool for validating \emph{other} metrics which \emph{can} be computed on real data.

If no single metric is completely satisfactory, it is often useful to measure both performance on a downstream task (which is practically relevant but indirect and expensive), as well as cheaper metrics which measure PPCs more directly.
For the downstream task, we consider \emph{Transductive active learning}~(TAL), which chooses training data interactively to quickly improve the performance at known testing locations. %
Since the test locations are known, a good strategy is to query points that have high PPCs with test locations~\citep{mackay1992information}. Hence, better PPCs ought to lead to better TAL performance. We validate the TAL task itself as a benchmark for PPCs, and find that the TAL performances closely align to metacorrelations on synthetic datasets where both can be evaluated. We also find that TAL algorithms make better use of uncertainty than ordinary active learning algorithms.

Because TAL requires training each model multiple times and averaging over many training runs, it is too expensive to use as a routine metric to guide algorithm development. Hence, we need a proxy which is cheaper to evaluate. We observe that predictive joint log-likelihoods are insufficient as a metric because they depend too heavily on the marginal uncertainty. To disentangle the effect of predictive correlations, we introduce \emph{cross-normalized log-likelihood (XLL)}, which measures the predictive joint likelihoods of a model after its marginal predictions have been transformed to match a reference model. We find that XLL closely aligns with both metacorrelations and TAL performance, validating its usefulness as a routine metric for improving PPC estimation.

Using the three metrics we developed, we evaluate a variety of Gaussian process and Bayesian neural net models in terms of PPC estimation. We first observe that these three metrics align well with each other on synthetic data. Further, under these metrics, we consistently find that different models differ significantly in the accuracy of their PPC estimations. %

%% file: tex/preliminary.tex
\section{Setup}\label{sec:setups}

\textbf{Notations.} Let $\vx \in \mathbb{R}^d$ be the input vector and $f$ be the function. We focus on the single-output regression problem  with Gaussian observation noise \citep{gal2016dropout, hernandez2015probabilistic, lakshminarayanan2017simple}, i.e., the target $y$ is a noisy observation of the function value, $y = f(\vx) + \epsilon, \epsilon \sim \normal(0, \sigma_{n}^2(\vx))$ \footnote{Following the original papers of each model, the observation variance $\sigma_{n}^2(\vx)$ is constant for all models other than Deep Ensemble~\citep{lakshminarayanan2017simple}. If Deep Ensemble uses a constant $\sigma_{n}^2(\vx)$, severe overfitting occurs in our experiments.}. We use $\mathcal{D}_{\mathrm{tr}}$ and $\mathcal{D}_{\mathrm{te}}$ to denote the training set and the test set, respectively. For active learning problems, we also use $\mathcal{D}_{\mathrm{pl}}$ to denote the pool set. Given the training set, for $n$ locations $\mathbf{X} \in \mathbb{R}^{n\times d}$, the predictive distribution for $f(\mathbf{X}) | \mathcal{D}_{\mathrm{pl}}$ is represented as a multivariate Gaussian $\mathcal{N}(\bm{\mu}(\mathbf{X}), \bm{\Sigma}(\mathbf{X}, \mathbf{X}))$. Here $\bm{\mu}(\mathbf{X}) = [\mu_{\vx}]_{\vx \in \mathbf{X}}$ is the predictive mean and $\bm{\Sigma}(\mathbf{X}, \mathbf{X}) = [\Sigma(\vx, \vx')]_{\vx,\vx' \in \mathbf{X}}$ is the predictive covariance. Then the predictive variance $\sigma^2_{\vx} = \Sigma(\vx, \vx)$ and the predictive correlation satisfy
\begin{align}
\rho(\vx, \vx')= \Sigma(\vx, \vx')/(\sigma_\vx \sigma_{\vx'}).
\end{align}

\subsection{Models}
We consider the following models in our experiments:

\textbf{GP/SVGP. } A Gaussian process (GP) defines a distribution $p(f)$ over functions $\mathcal{X} \to \mathcal{R}$ for some domain $\mathcal{X}$. For any finite set $\{\mathbf{x}_1, ..., \mathbf{x}_n\} \subset \mathcal{X}$, the function values $\mathbf{f} = (f(\mathbf{x}_1), f(\mathbf{x}_2), ..., f(\mathbf{x}_n))$ have a multivariate Gaussian distribution $\mathbf{f} \sim \mathcal{N}(\bmu, \mathbf{K}_{XX})$, where $\bmu:=[\mu(\bx_1), ..., \mu(\bx_n)]^{\top}$ is computed from the mean function $\mu(\cdot)$, and $\mathbf{K}_{XX}$ denotes the matrix $[k(\mathbf{x}_i, \mathbf{x}_j)]_{i,j}$ using the kernel function $k(\cdot, \cdot)$. For Gaussian likelihoods with variance $\sigma_n^2$, we can make predictions $p(y_*|\mathbf{x}_*, \mathcal{D}_{\mathrm{tr}})$ in closed form:
\begin{align}
 &p(y_*|\mathbf{x}_*, \mathcal{D}_{\mathrm{tr}}) = \mathcal{N} (\mathbf{K}_{*X}(\mathbf{K}_{XX} + \sigma_n^2\mathbf{I})^{-1}\mathbf{y}, \bm{\Sigma} ),\notag \\
 &\bm{\Sigma}  := \mathbf{K}_{**} - \mathbf{K}_{*X}(\mathbf{K}_{XX} + \sigma_n^2\mathbf{I})^{-1}\mathbf{K}_{X*}.+ \sigma_n^2. \notag
\end{align}
Exact posterior inference of GPs suffers from a computational cost that scales cubically with the number of training data. Thus, for large scale problems, Stochastic Variational Gaussian Processes (SVGPs) \citep{titsias2009variational, hensman2013gaussian} are usually adopted.

\textbf{Bayesian Neural Networks (BNNs). } Given an $L$-layer neural network, the weights $\bw=\{\bw_l\}_{l=1}^L$ are the collection of $V_l \times (V_{l-1}+1)$ weight matrices in each layer, where $V_l$ is the number of units in the $l$-th layer and the $+1$ accounts for the biases. Assuming the pre-activations and activations in the $l$-th layer are $\bz_l$ and $\ba_l$, we have $\ba_l = \bw_l^{\top} \bz_{l-1} / \sqrt{V_{l-1}+1}$, where $\sqrt{V_{l-1}+1}$ keeps the scales of $\ba_l$ independent of the number of input neurons. BNNs specify a prior distribution $p(\bw)$ over the weights and perform posterior inference for uncertainty estimation. In experiments, we use $p(\bw)=\mathcal{N}(\bm{0}, \eta \mathbb{I})$, where the scalar $\eta$ is the prior variance.

\textbf{HMC BNNs. }
The posterior over weights $p(\bw | \data)$ is intractable in BNNs. Hamiltonian Monte Carlo (HMC) \citep{neal2011mcmc} incorporates the score function and momentum in the proposals of Markov Chain Monte Carlo (MCMC), which makes it more efficient in high dimensional parameter spaces.
However, \citet{neal2011mcmc} points out that jointly sampling weights and the prior variance $\eta$ makes it difficult to mix; he suggests performing Gibbs sampling for prior variances combined with HMC for weights. Instead, we adopt Monte-Carlo Expectation-Maximization \citep{delyon1999convergence} for optimizing hyperparameters: the prior variance $\eta$ and the observation variance $\sigma_n^2$. Specifically, we run HMC for the weights and directly optimize hyperparameters using maximum-likelihood under the current weights.

\textbf{Bayes By Backprop (BBB). }
Variational methods provide another way to resolve the intractable posterior problem. They fit an approximate posterior $q(\bw)$ to maximize the evidence lower bound:
\begin{align}
\mathcal{L}_q = \mathbb{E}_{q}[\log p(\mathcal{D}_{\mathrm{tr}} | \bw)] - \KL{q(\bw)}{p(\bw)}. \notag
\end{align}
Specifically, Bayes-by-backprop (BBB)~\citep{blundell2015weight} uses a fully factorized Gaussian $\mathcal{N}(\bmu, \bm{\sigma}^2)$ to approximate the posterior.  Using the reparameterization trick~\citep{kingma2013auto}, the gradients towards $(\bmu, \bm{\sigma}^2)$ can be computed by backpropagation, and then be used for updates. We fix the prior variance $\eta=1$ because we did not find a benefit to optimizing it.

\textbf{Noisy Natural Gradients (NNG). }
Among the variational Bayesian families, the noisy natural gradient (NNG) \citep{zhang2017noisy} is an efficient method to fit multivariate Gaussian posteriors by adding adaptive weight noise to ordinary natural gradient updates. Assuming $q(\bw) = \mathcal{N}(\mean, \mSigma)$ and $\mathcal{D}\bw = \nabla_{\bw}\log p(y|\vx, \bw)$, then the update rules are:
\begin{align}%
    \mF &\leftarrow (1-\beta) \mF + \beta \left( \mathcal{D}\bw\mathcal{D}\bw^{\top} \right), \notag \\
    \mean &\leftarrow \mean + \alpha \left( \mF + \frac{1}{N \eta} \mI \right)^{-1} \left(\mathcal{D}\bw - \frac{1}{N \eta} \bw \right).\notag
\end{align}
In particular, when using a Kronecker-factored approximation~\citep{martens2015optimizing} for the Fisher matrix $\mF$, NNG is equivalent to imposing a matrix-variate Gaussian distribution for the variational posterior $q(\bw)$. Similarly to BBB, we fix the prior variance $\eta=1$ because we did not find a benefit to optimizing it.

\paragraph{Functional BNNs (FBNNs)} The weights usually have a complicated relationship with the corresponding function, making it difficult to specify a meaningful prior over weights. Functional BNNs \citep{sun2019functional} propose to directly apply priors over the function and perform functional variational inference. For estimating the functional KL divergence $\KL{q(f)}{p(f)}$, they use mini-batch marginal KL divergences,
\begin{align}
 \mathcal{L}_q = \mathbb{E}_{q}[\log p(\mathcal{D}_{\mathrm{tr}} | f)] - \lambda \KL{q\left( \begin{bmatrix}\vf^{\mathcal{D}_s} \\ \vf^M\end{bmatrix} \right)}{p\left(\begin{bmatrix}\vf^{\mathcal{D}_s} \\ \vf^M\end{bmatrix}\right)}, \notag 
\end{align}

where $\vf^{\mathcal{D}_s}$ and $\vf^M$ are the function values on locations within a random mini-batch $\mathcal{D}_s$ and on $M$ random locations from a heuristic distribution, respectively. Across all experiments, we set $\lambda=\abs{\mathcal{D}_\mathrm{tr}} / \abs{\mathcal{D}_s}$ and we use a GP-RBF prior. Following \citet{shi2019scalable}, we use a RBF random feature network \citep{rahimi2008random} with one hidden layer for the posterior $q(f)$, where the first layer is deterministically trainable and the second layer is a Bayesian linear regression layer applied on the features from the first layer. In consequence, the posterior distribution $q$ becomes Gaussian with explicit means and covariances. We set the heuristic distribution for $\vf^M$ as Gaussian distributions centered at random training locations with variances being $\mathrm{s}^2/2$, where $\mathrm{s}$ is the lengthscale of the prior kernel.

\textbf{Dropout BNNs. }
Dropout \citep{srivastava2014dropout} is a technique employed in training neural networks wherein each unit of each layer is possibly discarded independently with probability $p$. %
Typically, at test time, the trained network is made deterministic by scaling the outputs of each layer by $1-p$ rather than randomly zeroing them out. For obtaining uncertainty, \citet{gal2016dropout} propose to keep dropout stochastic at test time and use many forward passes to compute the means and variances. They show that dropout can be seen as implicitly optimizing a variational objective.

\textbf{Deep Ensemble. }
The Deep Ensemble \citep{lakshminarayanan2017simple}  consists of training multiple networks independently under the same objective. Then the predictions of all networks can be aggregated to compute the predictive means and variances.
Notably, the networks in the deep ensemble output not only the predicted mean $\mu_i(\bx)$, but also the predicted variance $\sigma^2_i(\bx)$. Each $\sigma^2_i(\bx)$ corresponds to the aleatoric variance; each $\mu_i(\bx)$ can be seen as a random function sample, so the epistemic uncertainty can be computed through the variance of $\{\mu_i(\bx)\}_{i=1}^m$.

\subsection{Covariance Computation} %
The posterior predictive correlations (PPCs) can be obtained from the predictive covariance $\bm{\Sigma}(\mathbf{X}, \mathbf{X})$ by $\rho(\vx, \vx')= \Sigma(\vx, \vx')/(\sigma_\vx \sigma_{\vx'})$. The covariance matrices of GP and FBNN are explicit. For example, for a GP with the kernel $K$, the predictive covariance given training inputs $\mathbf{X}_{tr}$ is, 
\begin{align}
K(\mathbf{X}, \mathbf{X}) - K(\mathbf{X}, \mathbf{X}_{tr})\left(K(\mathbf{X}_{tr}, \mathbf{X}_{tr})+\sigma_n^2 \bI \right)^{-1}K(\mathbf{X}_{tr}, \mathbf{X}), \notag
\end{align}
For other models such as BNNs, we need to draw samples to estimate the covariance matrix. For two points $\vx, \vx'$, we use a Monte Carlo estimate of the covariance:
\begin{align}
	\reallywidehat{{\Sigma}}(\vx, \vx') &= \frac{1}{m}\sum_{i=1}^{m}\left(f_i(\vx)-\reallywidehat{\mu}_\vx\right)\left(f_i(\vx')-\reallywidehat{\mu}_{\vx'}\right), \notag
\end{align}
where $f_i \sim p(f|\mathcal{D}_{\mathrm{tr}})$ are random function samples from the posterior, and $\reallywidehat{\mu}_\vx:=\frac{1}{m} \sum_{i=1}^m f_i(\bx)$. %

%% file: tex/method.tex
\section{Benchmarking PPC Estimators}

In this section, we describe our methodology for evaluating PPC estimators. We first introduce \emph{metacorrelations} as a gold standard when the true PPC values can be obtained.
Then, for practical settings, we introduce transductive active learning (TAL) as a downstream task for evaluating PPCs. Lastly, since TAL is expensive to run, we introduce \emph{Cross-Normalized Log Likelihood}~(XLL) as a more efficient proxy for TAL. 

\subsection{If We Have an Oracle Model: Metacorrelations}
The most conceptually straightforward way to evaluate PPC estimators is by comparing their estimates to the ground truth PPC values. When the data generating distribution is known and has a simple form (e.g.~synthetic data generated from a GP), we can compute ground truth PPCs; we refer to this as the \emph{oracle model}. We can then benchmark models based on how closely their PPCs match those of the oracle. We quantify this by computing the Pearson correlation of the PPCs of the candidate model with the PPCs of the oracle model; we refer to these correlations of correlations as \emph{metacorrelations}. Higher metacorrelations imply better PPC estimates.

\subsection{Evaluating PPC Estimators on Transductive Active Learning}\label{sec:acquisition_for_al_and_tal}

\begin{figure}[t]
\vspace{-0.4cm}
\centering
\includegraphics[width=0.42\textwidth]{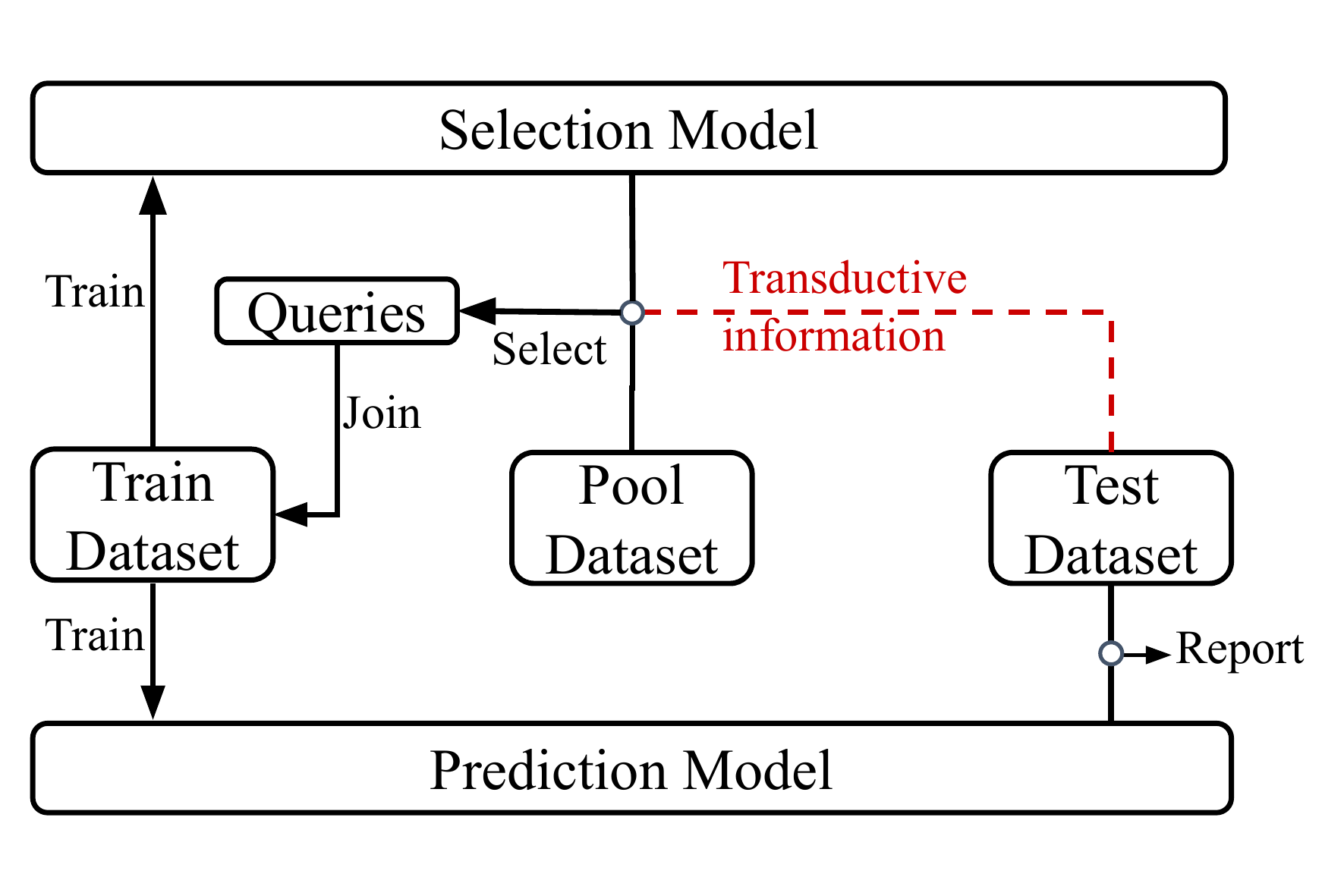}
\vspace{-0.5cm}
\caption{A diagram for the pipeline of ({\color{red}transductive}) active learning.\label{fig:al-diag}}
\vspace{-0.6cm}
\end{figure}

\emph{Active Learning} improves sample efficiency by allowing the learning algorithm to choose training data interactively. In each iteration, we use the \textit{selection model} to compute the \textit{acquisition function} for choosing points, and we report the test performance of the \textit{prediction model} trained on the accumulated training data. A diagram visualizing the active learning process is given in Figure~\ref{fig:al-diag}. For practical applications, the selection model is likely to be the same as the prediction model. However, we are interested in active learning as a downstream task for evaluating uncertainty estimates. Therefore, we fix a single prediction model across all conditions, and vary only the selection model, as this is the only part of the active learning algorithm that uses the uncertainty estimates.

The \textbf{Total Information Gain} (TIG)~\citep{mackay1992information}  is one acquisition function that has been used to benchmark predictive marginal uncertainty~\citep{cohn1996active, houlsby2011bayesian, hernandez2014predictive,zhang2017noisy, gal2017deep, beluch2018power}. 
TIG computes the informativeness of each point by measuring the mutual information between it and the model parameters $\vw$: $\mathrm{TIG}(\vx) \coloneqq  \mathbb{I}(y_{\bx}; \bw |\dataset_{\mathrm{tr}})$, where $\mathbb{I}$ represents mutual information. When the observation noise is assumed to be homoscedastic, choosing points by TIG is equivalent to choosing the points with the largest predictive variances. However, as pointed out by~\citet{mackay1992information}, TIG is problematic since it favors points a at the edges of the input space, hence leading to useless or even harmful queries. %
Moreover, as TIG relies only on marginal uncertainty, it cannot be used for benchmarking PPCs.

Instead we consider the \emph{Transductive Active Learning}~(TAL) setting for benchmarking PPCs. In TAL, a region of interest, such as the locations of test points or the distribution of the locations, is assumed to be known in advance. %
\citet{mackay1992information} presents the \textbf{Marginal Information Gain} (MIG) acquisition function, which measures the information gained about the region of interest by querying a new point. Supposing $\vx_u$ is the point of interest\footnote{For a test set $\{\vx^i_u\}_i$ of interest, we adopt the mean marginal information gain~\citep{mackay1992information}.}, $\mathrm{MIG}(\vx;\vx_{u}) \coloneqq  \mathbb{I}(y_{\bx}; f(\bx_u) |\dataset_{\mathrm{tr}})$. We prefer MIG to TIG both because it's been found to avoid TIG's outlier problem \citep{mackay1992information} and because it makes direct use of PPCs. %

In practice, selecting one point at a time, and re-training the models in between, is unreasonably expensive, so instead we would like to select batches of points. Na{\"\i}vely selecting the set of points with highest scores often results in inefficiently selecting a cluster of nearby and redundant points.
To encourage diversity in a batch, we introduce the \textbf{Batch Marginal Information Gain} (BatchMIG) acquisition function, inspired by the BatchBALD algorithm~\citep{kirsch2019batchbald} in active learning. For a batch $\bx_{1:q}$, 
\begin{align}
\label{eq:BMIG}	
    &\textrm{BatchMIG}(\vx_{1:q}; \vx_u)\coloneqq \mathbb{I}(\vy_{\bx_{1:q}}; f(\bx_u) |\dataset_{\mathrm{tr}}). %
\end{align}
BatchMIG quantifies the amount of information carried by the selected batch.
Though selecting the optimal batch $\vx_{1:q}^{\star}$ for BatchMIG is intractable, we adopt a greedy algorithm \citep{kirsch2019batchbald} to approximate it. We note that BatchMIG exploits PPCs more fully than MIG: in addition to using transductive information, it also uses PPCs between candidate query points to encourage diversity. Hence, it appears especially well-suited for benchmarking PPC estimators.

\begin{figure}[t]
\vspace{-0.4cm}
\hspace{-0.8cm}
\includegraphics[width=0.55\textwidth]{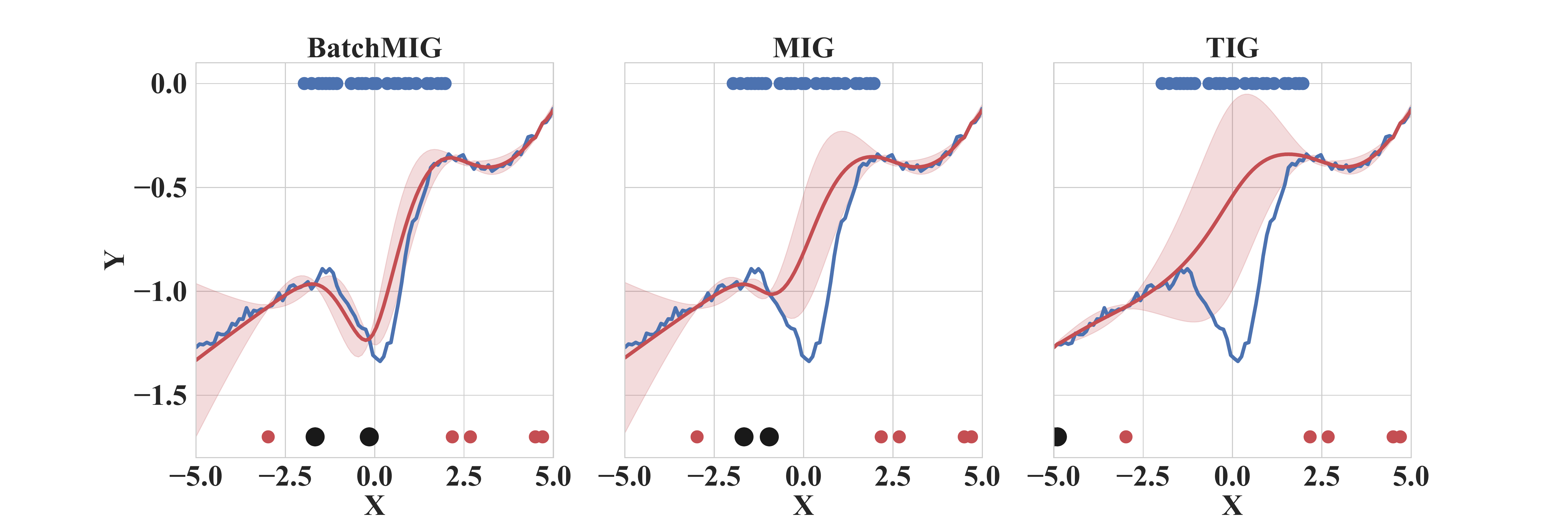}
\vspace{-0.5cm}
\caption{A visual comparison between BatchMIG, MIG and TIG. The {\color{red}red curve} is the prediction after query, {\color{blue}blue curve} is the ground-truth, the {\color{blue}$\bullet $} on top are the testing points, the {\color{red}$\bullet$} are the training points, and the $\bullet$ are the queried points of the corresponding acquisition function.\label{fig:acquisition-comparisons}}
\vspace{-0.45cm}
\end{figure}

To illustrate the differences between acquisition functions, we compare TIG, MIG and BatchMIG through a toy example shown in Figure~\ref{fig:acquisition-comparisons}. We observe that TIG chooses points nearby the boundary and MIG chooses redundant points close to each other, while BatchMIG chooses diverse points close to the test set. In consequence, the BatchMIG predictions after query match the ground truth better at the test locations. %

\subsection{Efficient Metrics beyond TAL}
\label{sec:novel_metrics}
TAL is a useful downstream task for evaluating the PPCs, but running TAL requires training the model multiple times in succession, which is computationally expensive. Furthermore, as the PPCs are only part of a larger pipeline, it is rather indirect as a means of evaluating PPC estimators. Both factors limit its routine use as a guide for algorithmic development. In this section we introduce more direct and efficient metrics for PPC evaluation.

\subsubsection{Joint Log-Likelihoods?}%
\label{subsec:issue_joint_lld}

The log marginal likelihood $\sum_{i=1}^n \log p(y_i| \bx_i)$ is a widely used metric for evaluating predictive marginal uncertainty. By analogy, it would be natural to evaluate PPCs using the \emph{joint} log-likelihood for a batch of points, $\log p(y_1, ..., y_b | \bx_1, ..., \bx_b)$. However, this is unsatisfying for two reasons:

\textbf{Impact of predictive marginals.} We've found the joint log-likelihood scores to be determined almost entirely by the marginal log-likelihood scores, with only a small dependence on the PPCs. Hence, in practice, they provide little new information beyond marginal log-likelihoods, as shown by Figure~\ref{fig:app:marginal_joint} in the appendix.

\textbf{Uncorrelated random batches.} The points in a random batch $\{(\bx_i, y_i)\}_{i=1}^b$ are almost uncorrelated because they usually scatter far away from each other.

For both reasons, joint log-likehoods for random batches do not indicate the quality of PPCs. %

\subsubsection{Cross-Normalized Log Likelihood}
\label{subsec:practical_corr_eval}

As discussed in Section~\ref{subsec:issue_joint_lld}, joint log-likelihoods are appealing because they directly measure uncertainty and are efficient to evaluate, but they have the flaw that the scores are dominated by the marginal predictive distributions.
To disentangle the effect of predictive marginals, we propose to substitute each model's predictive means and variances with those of a \emph{reference model}, whose predictive means and variances are believed to be reasonably good. Consequently, the joint likelihoods depend only on the PPCs but not on the predictive marginals. We refer to this criterion as \emph{cross-normalized log-likelihood (XLL)}. To compute the XLL of a batch of $b$ points $\{\mX, \vy\}$, we define
\begin{align}
	&\mathrm{XLL}(\vy|\mX, \mathcal{M}, \mathcal{M}_{\text{ref}}) \notag \\
	&= \log \mathcal{N}(\vy| \vmu_{\text{ref}},\; \diag(\bsigma_{\text{ref}}) \mC_{\mathcal{M}} \diag(\bsigma_{\text{ref}}))	.
\end{align}
Here, $\mathcal{M}$ and $\mathcal{M}_{\text{ref}}$ denote the candidate and reference model, $\vmu_{\text{ref}}$ and $\bsigma_{\text{ref}}^2$ are the predictive mean and variance given by $\mathcal{M}_{\text{ref}}$, and $\mC_\mathcal{M}$ is the predictive correlation matrix computed by $\mathcal{M}$. Additionally, we can also rank each candidate model by the XLL under the same reference model, and we refer to the resulting criterion as \emph{cross-normalized log-likelihood rank (XLLR)}. Furthermore, to mitigate the problem that most pairs of randomly chosen locations have small correlations between the function values, we use the reference model to select top-correlated points (sorted by the absolute value of correlations) to form batches for evaluating the joint log-likelihoods.

\textbf{Choosing the reference model.} Intuitively, the ideal reference model would be the oracle, i.e. the true data generating distribution. That way, the XLL fully represents how the PPCs $\mC_{\mathcal{M}}$ match the ground truth correlations  $\mC_\text{gt}$. Although the predictive marginals of the oracle model are unknown in real-world problems, we can show that, as long as the reference marginals are nearly optimal, the XLL still reflects the quality of the PPCs. Informally, let $\mathrm{LogDet}(\mC_\text{gt}, \mC_{\mathcal{M}})$ denote the LogDet divergence \citep{kulis2006learning}; then,
\begin{align}
\mathrm{LogDet}(\mC_\text{gt}, \mC_{\mathcal{M}}) = - \expect_{\vy|\mX} \mathrm{XLL} + \mathcal{O}\left(\frac{b^{3/2}}{\lambda}\sqrt{\xi}  \right)+ c , \notag 
\end{align}
where $c$ is a constant, $b$ is the batch size (set to $5$ in our experiments). Here, $\xi$ denotes the KL divergence between the reference marginal $(\vmu_{\text{ref}}, \bsigma_{\text{ref}})$ and the true marginal $(\vmu_{\text{gt}}, \bsigma_{\text{gt}})$. $\lambda$ denotes the smallest eigenvalue of $\bC_{\mathcal{M}}$, which is usually not much smaller than $1$ due to the observation noise. This results indicates that, with a nearly-optimal reference marginal, a larger XLL implies a smaller LogDet divergence. A more formal statement is given as Theorem~\ref{thm:corr-est} in the appendix.

To avoid favoring any particular model in our comparisons, we propose to iterate through every candidate model to serve once as the reference model. The reported XLL and XLLR values are averaged over all choices of the reference model. Still, one would like to validate that the results are not overly sensitive to the choice of reference model. In Figure~\ref{fig:XLL-best}, we observe that the XLLR values are consistent between choices of reference model (see Figure~\ref{fig:XLL-best}). We also observe that XLL and XLLR align well with the TAL performances as well as the oracle-based metacorrelations. Pseudocode for computing XLL and XLLR can be found in Algorithm~\ref{alg:new-metric} in the appendix.

%% file: tex/related_works.tex
\section{Related Works}
\textbf{Benchmarking Uncertainty.} There have been numerous attempts to reliably evaluate Bayesian models. The UCI regression benchmarks \citep{hernandez2014predictive} are used extensively for evaluating Bayesian neural networks. Calibration metrics \citep{guo2017calibration, kuleshov2018accurate} are used for testing whether the predictions are over- or under-confident. \citet{snoek2019can} studied how the predictive accuracy and calibration are affected by a dataset shift.
Researchers have also related the performance of various downstream tasks to the handling of uncertainty. \citet{riquelme2018deep} developed a contextual bandit benchmark which uses marginal uncertainty to balance exploitation and exploration. Diabetic retinopathy diagnosis \citep{oatml2019bdlb} was also used for comparing uncertainty estimates. However, all of these benchmarks are on evaluating marginal uncertainty estimations.

\textbf{Algorithms exploiting PPCs.} Although less attention has been paid to PPCs, there are still several algorithms that exploit PPCs in their designs. In transductive active learning \citep{mackay1992information, yu2006active}, mutual information gain (MIG) improves the data efficiency by gathering points that have high PPCs with the test set. Similarly in Bayesian optimization, entropy search approaches \citep{hennig2012entropy, hernandez2014predictive, wang2017max} make the query to acquire the most information about the optimum of the underlying function. Knowledge gradient \citep{frazier2009knowledge} makes the query so that the expected next step performance is maximized. Furthermore, in cost-sensitive Bayesian optimization where different queries might incur different costs,  obtaining cheap queries taht can indirectly  acquire information about more expensive ones~\citep{hennig2012entropy, swersky2013multi}. Nevertheless, they only concern the usage of PPCs, but leave the question of which model is better at predicting PPCs unanswered.

%% file: tex/active_learning.tex
\section{Experiments}
\label{sec:exps}
In this section, we first introduce the details of the setup in our experiments. Then, we conduct experiments on synthetic datasets to validate TAL and XLL(R) as metrics for evaluating PPCs. We show that both metrics correspond closely with our ground-truth metric, metacorrelations. Finally, we use our TAL and XLL(R) benchmarks to address the main question of this paper: how accurately can Bayesian regression models estimate PPCs in real-world datasets?

\subsection{Experimental Setup}

\textbf{Synthetic Datasets. } Some of our experiments were done on synthetic data drawn from a Gaussian process. Our motivation for this was twofold. Firstly, having access to the true data generating distribution allows us to compute metacorrelations with the oracle model. Secondly, the prior distributions for all models could be chosen to exactly or approximately match the true distribution.

We generated synthetic datasets using a Gaussian process whose kernel was obtained from the limiting distribution of infinitely wide Bayesian ReLU networks with one hidden layer~\citep{neal1995bayesian}. Hence, the priors for the finite BNN weights could be chosen to approximately match this distribution. To generate a $d$-dimensional dataset, we sampled $5d, 500, 200$ points from the standard Normal distribution, as the training, test and pool sets, respectively. Then we sampled a random function $f$ from the oracle Gaussian process. The corresponding observed function value at $\bx$ is then $y = f(\bx) + \epsilon, \epsilon \sim \mathcal{N}(0, 0.01)$. For all models, we used 1,000 epochs for training and the true observation variance\footnote{For the synthetic setting, each network in the Deep Ensemble only predicts the mean.}. All results were averaged over 50 datasets which are randomly sampled in this manner.

\begin{figure}[t]
\vspace{-0.3cm}
    \centering
     \includegraphics[width=0.45\textwidth]{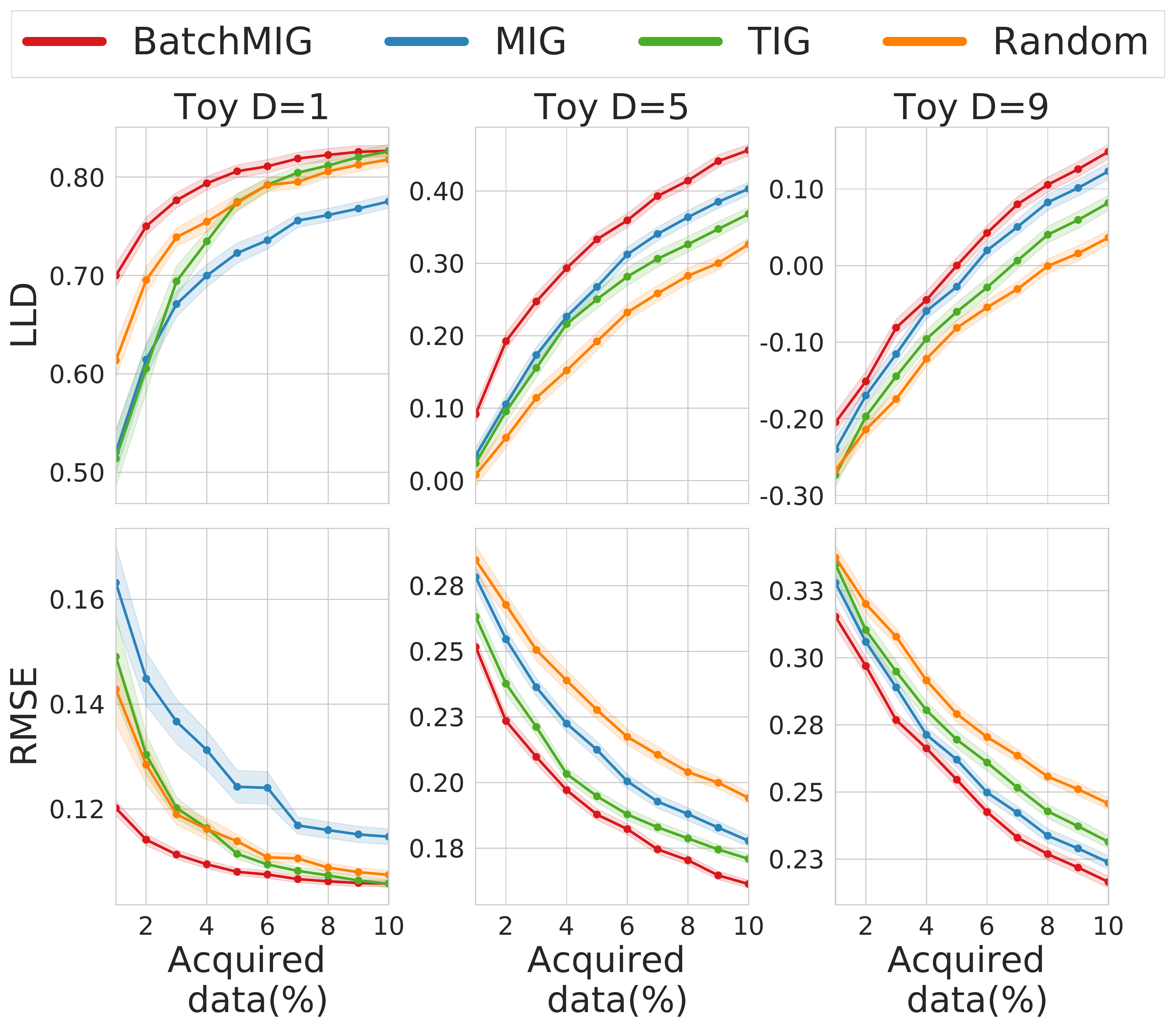}
    \caption{Results on synthetic datasets using BatchMIG, MIG, TIG and random selection with the oracle model.\label{fig:compare_criterion} }
    \vspace{-0.5cm}
\end{figure}

\textbf{UCI Regression Datasets. } We also conducted experiments using eight UCI regression datasets which have served as standard benchmarks for BNNs~\citep{hernandez2014predictive,gal2016dropout,zhang2017noisy,lakshminarayanan2017simple,sun2019functional}. This includes five small datasets~(\texttt{Boston}, \texttt{Concrete}, \texttt{Energy}, \texttt{Wine} and \texttt{Yacht}) and three large datasets~(\texttt{Kin8nm}, \texttt{Naval} and \texttt{Power\_Plant}). For all experiments, we used $20\%$ of the entire dataset as the initial training set $\mathcal{D}_{\mathrm{tr}}$, $20\%$ as the test set $\mathcal{D}_{\mathrm{te}}$, and the remaining $60\%$ as the pool set $\mathcal{D}_{\mathrm{pl}}$ in active learning. In each active learning iteration, we selected a number of points from the pool set corresponding to $1\%$ of the original dataset, so that the final prediction was made with $30\%$ of the data~(see Algorithm~\ref{alg:active-learning} in the Appendix). All experiments were run ten times with varied random seeds.

\textbf{An Oracle Prediction Model for AL. } The active learning performance is affected by two factors: the prediction model and the selection model. %
To disentangle their effects, we used the same  `Oracle' prediction model when varying the selection models. For the synthetic setting, we adopt the oracle GP model as the prediction model. For real-world data, the data generating distribution is unknown, so instead we fit the hyperparameters of a flexible GP kernel structure using the union of the training and pool set (rather than the much smaller initial training set). While the resulting distribution is probably not an exact match to the data generating distribution, it is likely superior to what the other methods can realistically obtain using the much smaller training set. For the flexible GP prior, we used the neural kernel network (NKN) kernel \citep{sun2018differentiable}, a differentiable architecture representing sums and products of simpler kernels.

\subsection{Is TAL a Suitable Downstream Task?} \label{subsec:trans-active}

Active learning using TIG has been used as a benchmark for uncertainty in BNNs, but the validity of this benchmark is debatable, as improved uncertainty modeling has not been shown to consistently benefit active learning performance~\citep{hernandez2014predictive}. We now aim to determine if \emph{transductive} active learning is a good downstream task for evaluating PPCs.

\textbf{Is transductive information useful for active learning? } We first conducted experiments on the synthetic datasets and UCI regression datasets to determine if transductive information is useful for active learning. Specifically, for the synthetic setting, we compared the active learning acquisition functions with all posterior predictive distributions obtained from the true GP prior that used to generate the data. For the realistic setting, we used the predictive distributions given by the `Oracle' NKN model for computing these criteria.
The results on synthetic datasets are reported in Figure~\ref{fig:compare_criterion} comparing TIG, MIG, BatchMIG and random selection. We first observe that the transductive criteria can in general achieve much better sample efficiency than the non-transductive ones. Furthermore, BatchMIG outperforms MIG, due to the increased diversity of points within the query batch. Both observations also hold on the UCI datasets; results can be found in~Figure~\ref{app:fig:compare_criterion_uci} (deferred to the Appendix to save space). Hence, it appears that transductive information can be exploited for sample efficiency.

\begin{figure}[t]
    \centering
\includegraphics[width=0.42\textwidth]{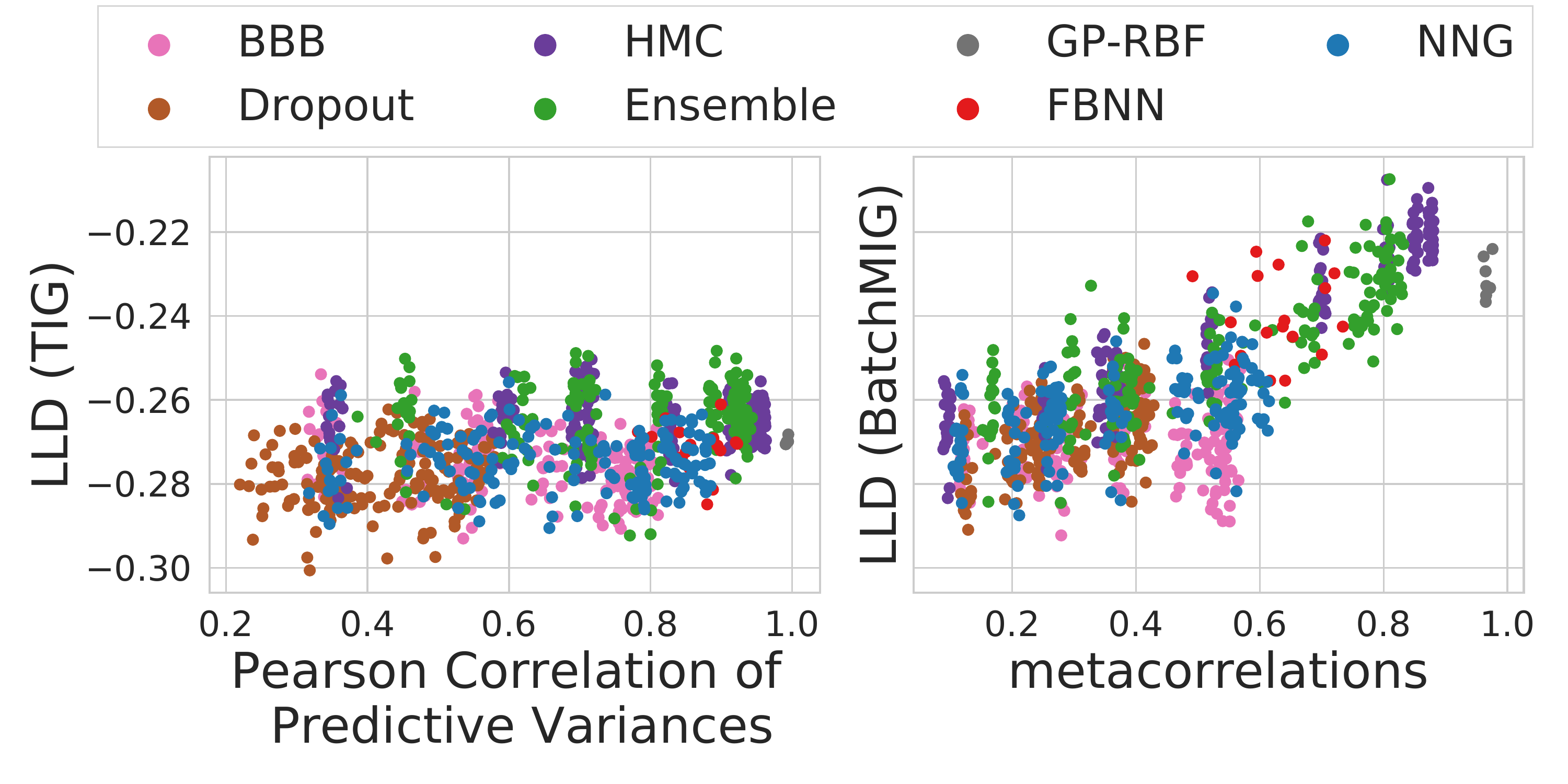}  
\vspace{-0.1cm}
\caption{Left: Active learning. TIG performance vs. the quality of predictive variances. Right: Transductive active learning. BatchMIG performance vs. the quality of predictive correlations (metacorrelations). The BatchMIG performance and the metacorrelations are strongly correlated with a coefficient \textbf{0.762}.\label{fig:scatters_lld_corr_var}}
\vspace{-0.6cm}
\end{figure}

\begin{table*}[t]
\vspace{-0.4cm}
\centering
\caption{ The average XLLR for each model on UCI datasets. We use color {\color{red}red} to highlight the best ones (within one standard error), and color {\color{blue}blue}  for the worst ones (within one standard error).}
\label{table:XN-LLDR}
\vspace{-0.2cm}
\resizebox{0.92\textwidth}{!}{
\begin{tabular}{lccccccc}
\toprule
\textbf{Dataset}/\textbf{Method} 
& (SV)GP-RBF & BBB & NNG & HMC & FBNN & Dropout & Ensemble
\\

\midrule
 \texttt{Boston}
&2.53 (0.20)
&4.24 (0.18)
&2.57 (0.19)
&\color{red}{0.93 (0.13)}
&3.20 (0.21)
&\color{blue}{5.31 (0.13)}
&2.21 (0.19)
\\

 \texttt{Concrete}
&2.13 (0.14)
&4.50 (0.14)
&3.19 (0.17)
&\color{red}{1.49 (0.18)}
&2.49 (0.14)
&\color{blue}{5.81 (0.09)}
&\color{red}{1.40 (0.21)}
\\

 \texttt{Energy}
&1.93 (0.17)
&4.07 (0.19)
&\color{blue}{4.71 (0.15)}
&\color{red}{1.59 (0.22)}
&2.00 (0.17)
&\color{blue}{4.57 (0.15)}
&2.13 (0.21)
\\

 \texttt{Wine}
&1.89 (0.18)
&4.61 (0.13)
&3.00 (0.16)
&2.14 (0.15)
&\color{red}{1.14 (0.21)}
&\color{blue}{5.59 (0.12)}
&2.63 (0.18)
\\

 \texttt{Yacht}
&\color{red}{1.90 (0.19)}
&3.74 (0.14)
&3.49 (0.17)
&2.63 (0.27)
&\color{red}{1.76 (0.16)}
&\color{blue}{5.71 (0.11)}
&\color{red}{1.77 (0.15)}
\\

\texttt{Kin8nm}
&\color{red}{1.24 (0.11)}
&4.37 (0.07)
&4.27 (0.16)
&1.44 (0.15)
&1.81 (0.15)
&\color{blue}{5.14 (0.25)}
&2.71 (0.18)
\\
 \texttt{Naval}
&3.33 (0.14)
&\color{blue}{5.54 (0.12)}
&4.76 (0.08)
&2.40 (0.18)
&2.31 (0.07)
&1.43 (0.28)
&\color{red}{1.23 (0.12)}
\\
 \texttt{Power\_plant}
&\color{red}{1.19 (0.11)}
&3.81 (0.12)
&\color{blue}{5.04 (0.15)}
&1.93 (0.15)
&3.54 (0.11)
&4.33 (0.29)
&\color{red}{1.16 (0.15)}
\\
\midrule
\textbf{Mean~$\downarrow$}
&2.02 
&4.36 
&3.88 
&\color{red}{1.82}
&2.28 
&\color{blue}{4.73}
&1.91
\\
\bottomrule

\end{tabular}
}
\vspace{-0.4cm}
\end{table*}

\textbf{Do more accurate PPCs enable better TAL performance?} To study this question, we used the synthetic data so that ground-truth correlations and variances (from the Oracle) were available. We conducted active learning experiments using various models for query selection, each with multiple choices of hyperparameters. For each model, we evaluate the test log-likelihoods after one iteration. From Figure~\ref{fig:scatters_lld_corr_var}, we find that the BatchMIG performance is well aligned with the quality of PPCs, as measured by the metacorrelations. Hence, TAL is likely to be a good downstream task for evaluating PPCs. In contrast, the TIG performance appears to be unrelated to the quality of predictive variances, as expected. The contrast between (BatchMIG, Correlation) and (TIG, Variance) highlights the usefulness of TAL for benchmarking PPCs. %

\begin{figure}[t]
    \centering
    \vspace{-0.3cm}
\includegraphics[width=0.43\textwidth]{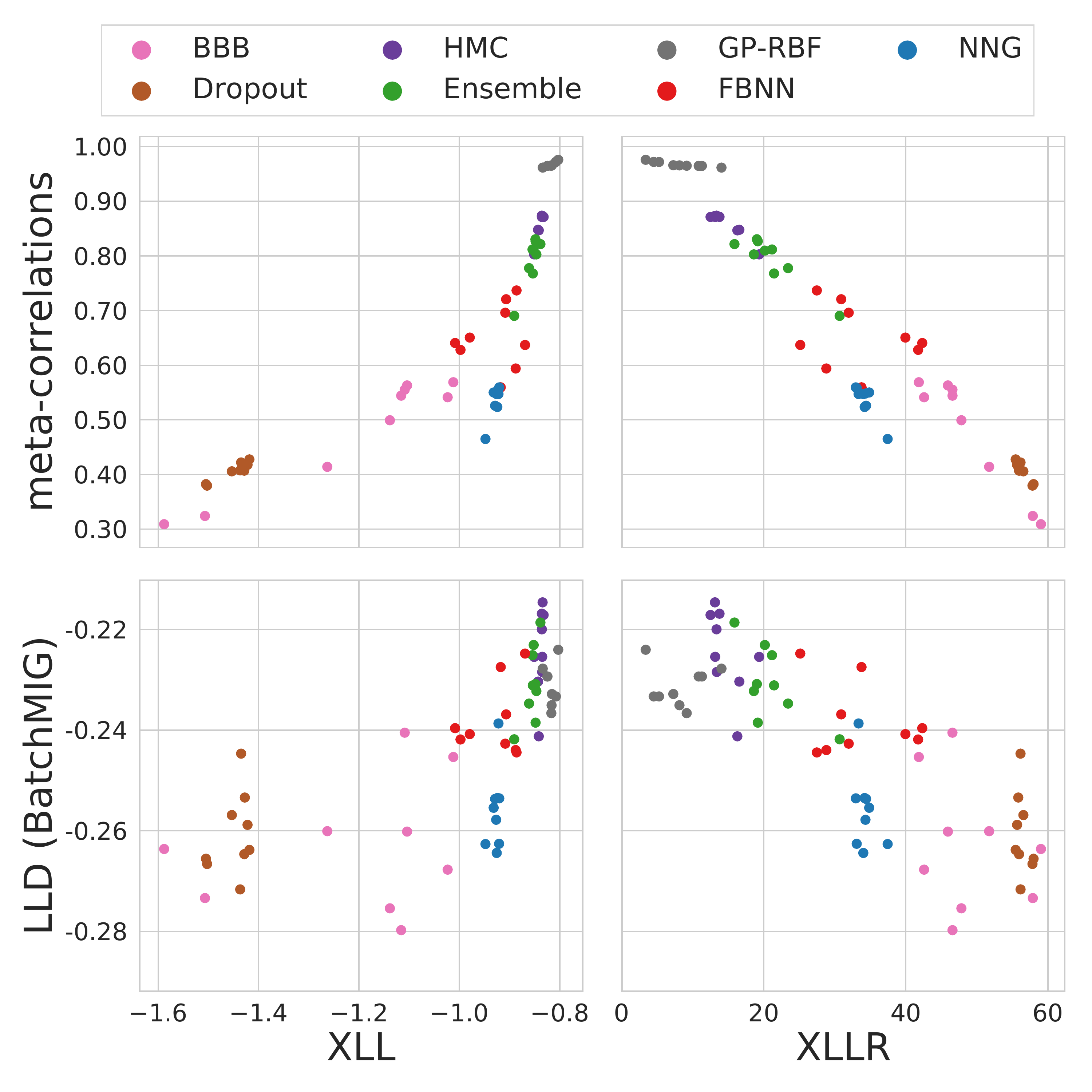}  
\vspace{-0.4cm}
     \caption{Metacorrelations and BatchMIG performance versus XLL and XLLR.\label{fig:exp_metrics_non_ideal}}
     \vspace{-0.7cm}
\end{figure}

So far, we have demonstrated that (1) we can enjoy significantly better sample efficiency in active learning by incorporating the transductive information; and (2) the TAL performance is clearly and positively correlated with the quality of the PPCs. These evidence support that TAL is a suitable benchmark for evaluating PPCs.

\subsection{Are XLL and XLLR Reliable?}

We have discussed three evaluation metrics with different strengths and weaknesses: TAL, which is meaningful but expensive to evaluate; metacorrelation, which is fast and directly measures correlations but requires an oracle model; and XLL/XLLR, which are fast and do not require an oracle, but are less obviously related to the quality of PPCs. How well do they match each other?  We measure all three metrics on the synthetic datasets. The results are shown in Figure~\ref{fig:exp_metrics_non_ideal}. We  observe that XLL/XLLR align well with both metacorrelations and the BatchMIG performance in TAL. This indicates that XLL/XLLR could be good proxy metrics for evaluating PPCs when TAL experiments are too expensive and the oracle model is not available.

\begin{figure}[t]
    \centering
    \vspace{-0.1cm}
    \includegraphics[width=0.43\textwidth]{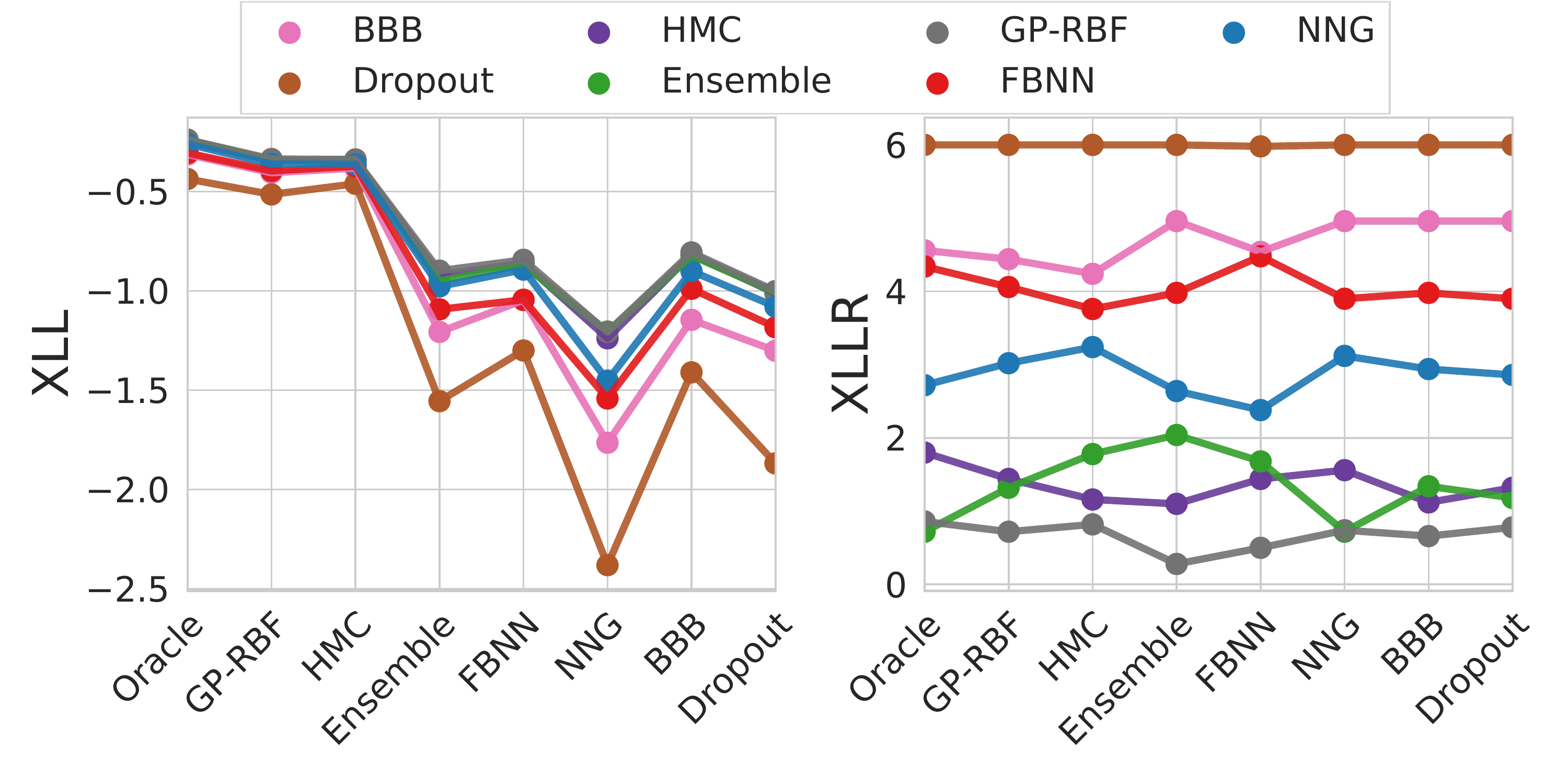}
    \vspace{-0.3cm}
    \caption{XLL and XLLR comparisons for each individual reference model. The x-axis indicates the reference model.\label{fig:XLL-best}}
    \vspace{-0.6cm}
\end{figure}

In Figure~\ref{fig:XLL-best}, we plot the XLL and XLLR for each individual reference model. The absolute XLL values depend on the reference model. However, we observe a relatively consistent ordering of different PPC estimators regardless of the reference model, suggesting that XLL and XLLR are relatively stable criteria for evaluating PPC estimators.

\begin{figure*}[t]
\vspace{-0.3cm}
    \centering
    \begin{subfigure}{0.49\textwidth}
    \includegraphics[width=\textwidth]{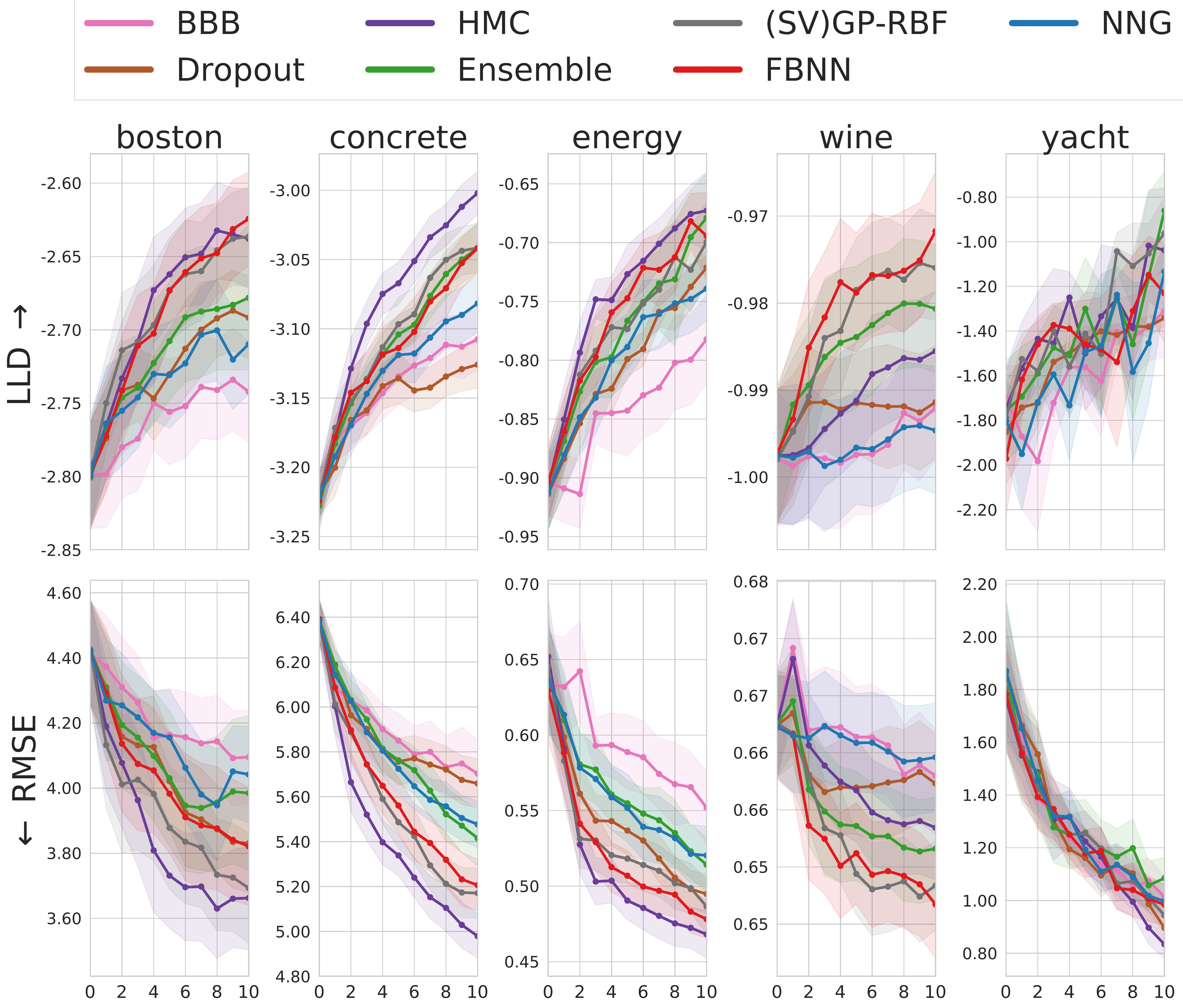}
    
    \includegraphics[width=\textwidth]{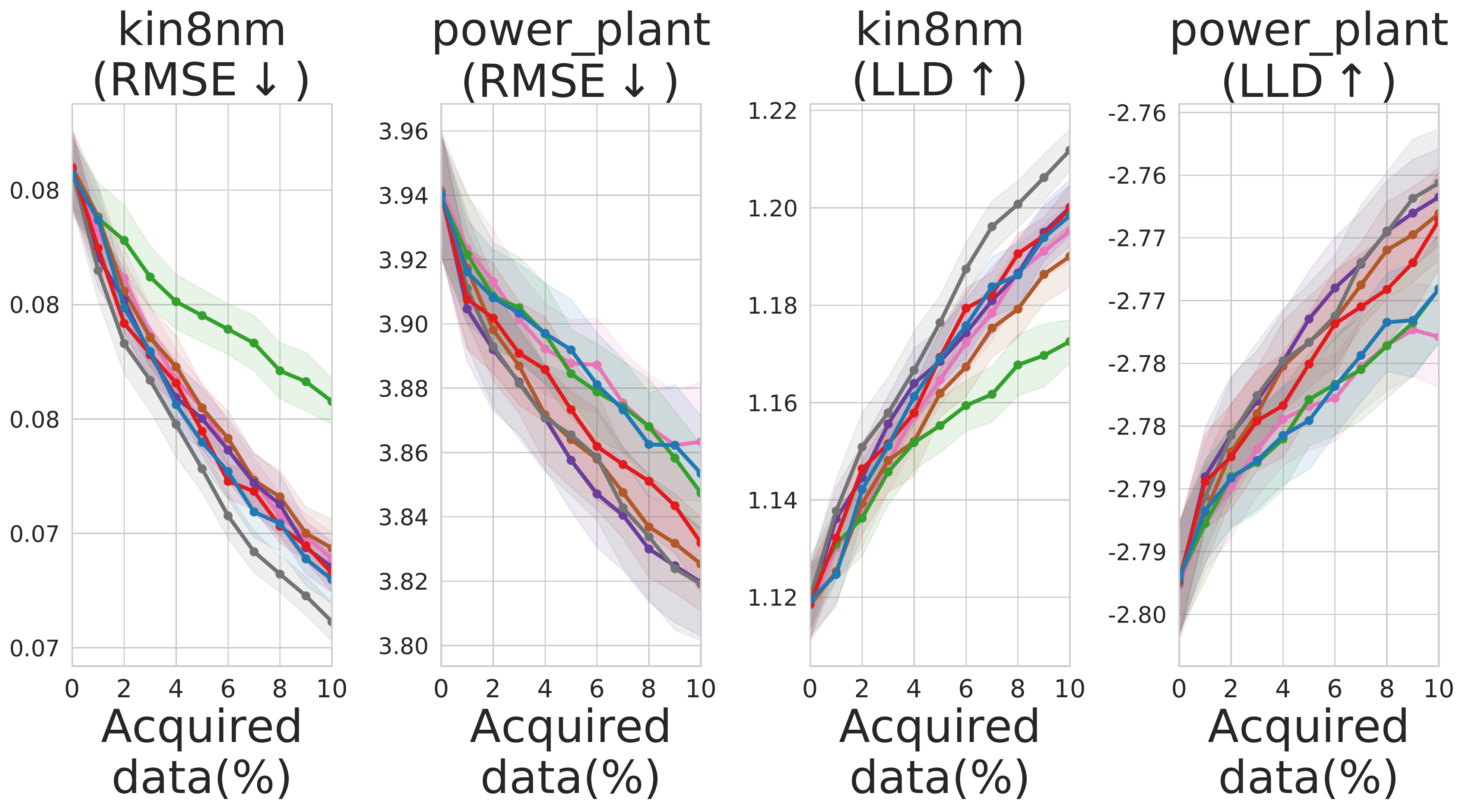}
    \caption{HMC BNN as the prediction model.\label{fig:active_learning_MMIG_of_each_model_HMC}}
    \end{subfigure}
    \hfill
    \hfill
    \begin{subfigure}{0.49\textwidth}
    \includegraphics[width=\textwidth]{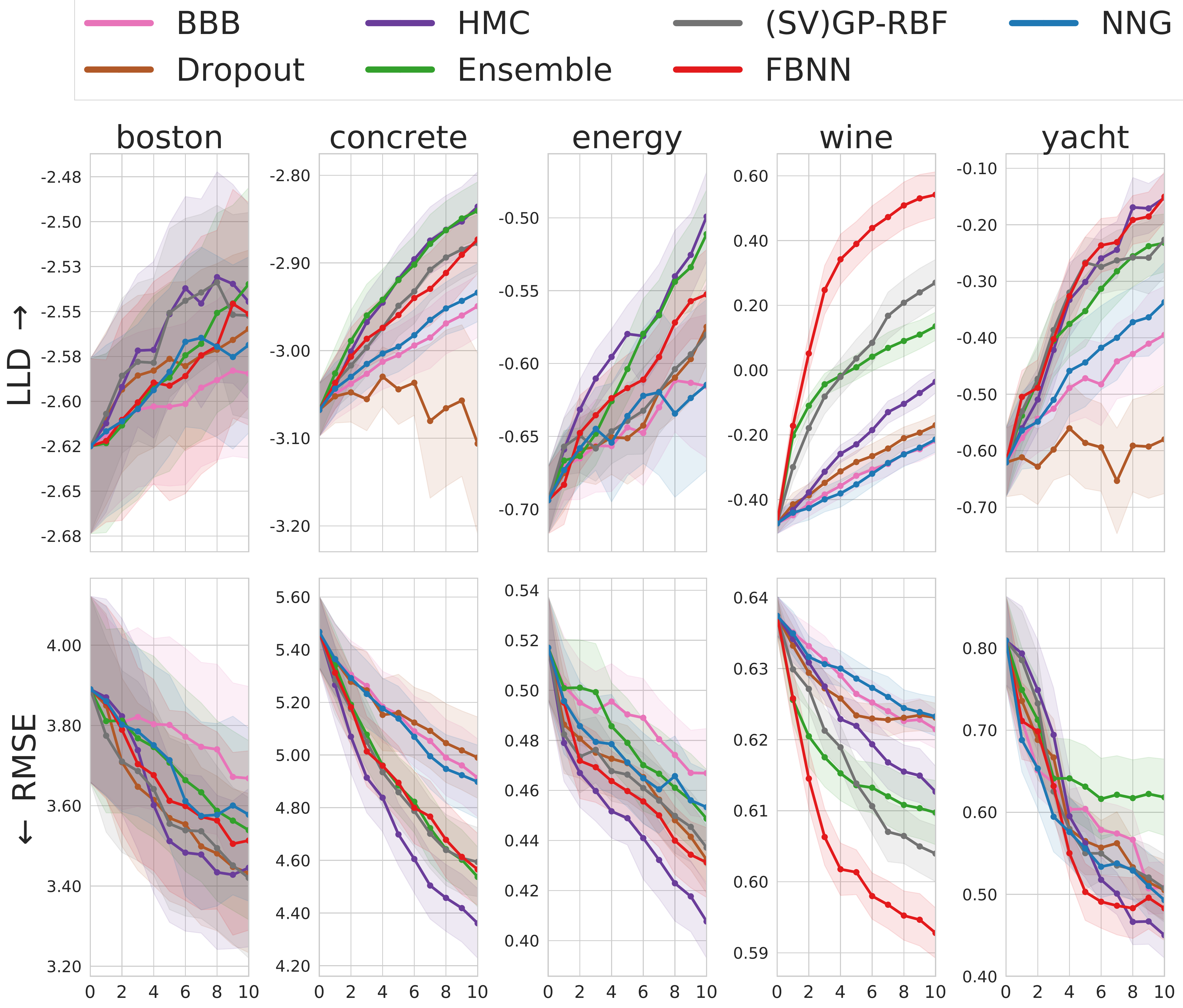}
    \
    \includegraphics[width=\textwidth]{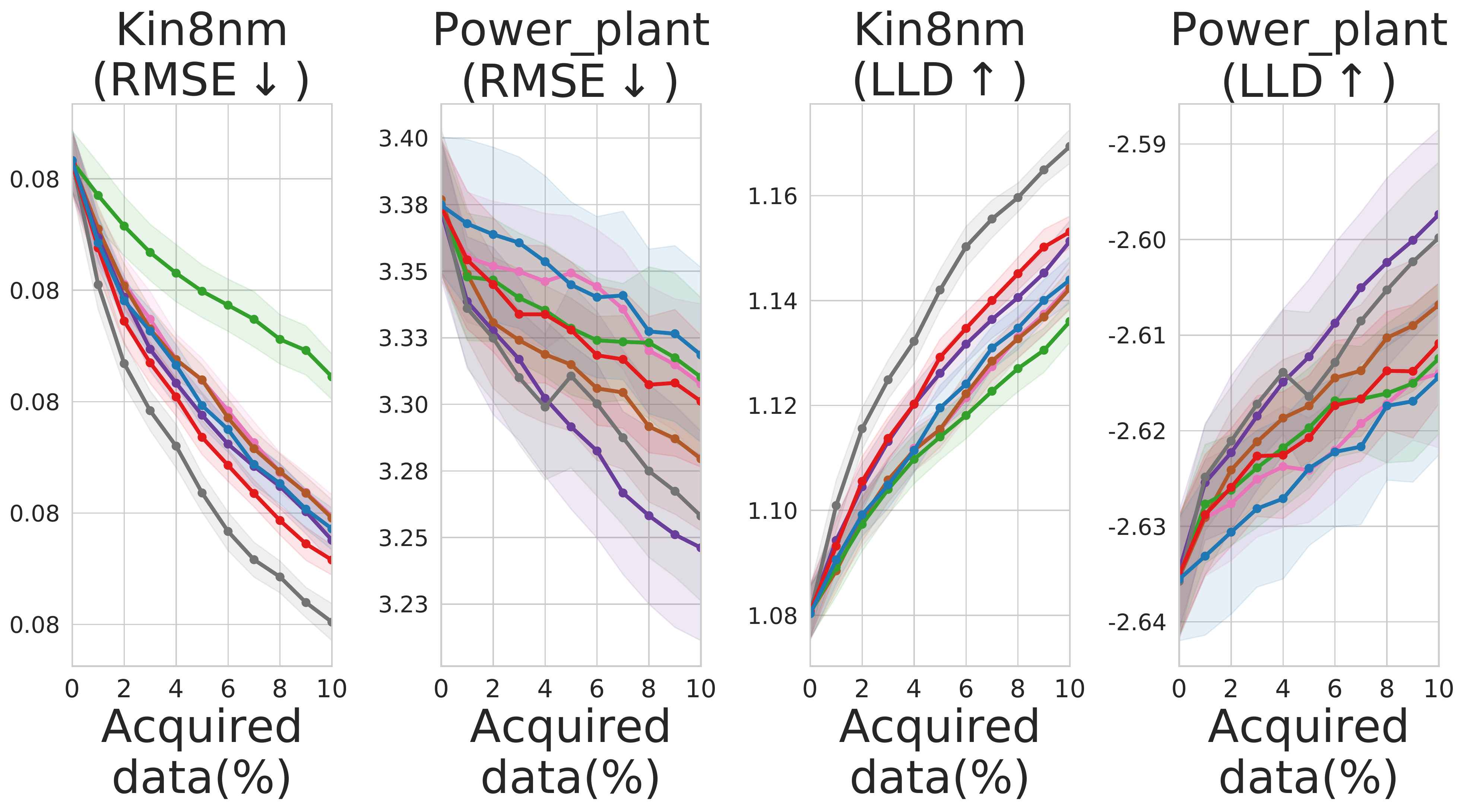}
    \caption{NKN as the prediction model. \label{fig:active_learning_MMIG_of_each_model}}

    \end{subfigure}
    \caption{BatchMIG-TAL using different selection methods, with (a) HMC BNN and (b) NKN as the prediction model.}
    \vspace{-0.4cm}
\end{figure*}

\subsection{PPC Estimation for Real-World Datasets}

The previous sections validated TAL and XLL/XLLR as metrics for the quality of PPCs. We now turn to our central question: how well can existing Bayesian models estimate PPCs?  For this section, we consider both XXL/XLLR and TAL on UCI datasets. We first talk about the empirical results on both benchmarks, and then discuss how do they connect to each other.

\textbf{XLL and XLLR. }
Since the XLL cannot be directly averaged across different datasets, we instead average the ranks (i.e.~XLLR values). This is analogous to the approach taken in the Bayesian bandits benchmark paper~\citep{riquelme2018deep}. We present the empirical results of XLLR computed for different models in Table~\ref{table:XN-LLDR}; results for XLL are given in Appendix~\ref{app:XN-LLD}.
 Firstly, we observe that HMC, GP-RBF, FBNN and Ensemble usually outperform BBB, NNG and Dropout.
 Moreover, GP-RBF and FBNN both conduct function-space inference under the same prior, thus they perform similarly with FBNN being slightly worse. 
Besides, in contrast to HMC, BBB and NNG cannot estimate the correlations well, which highlights the importance of accurate posterior inference in BNNs. Finally, we found that Dropout estimates the PPCs the worst overall, even though we observed that Dropout performs well in terms of log marginal likelihoods. %

\textbf{Transductive Active Learning. }
Figures~\ref{fig:active_learning_MMIG_of_each_model_HMC} and \ref{fig:active_learning_MMIG_of_each_model} show the TAL performance where various algorithms are used for query selection but predictions are made using the HMC and NKN models, respectively. (We omitted \texttt{naval}, since active learning appears to provide no benefit; see Figure~\ref{app:fig:compare_criterion_uci} in the appendix). Clearly, the model used for query selection has a large effect even when the prediction model is fixed. In general, even though the prediction models are different, the best-performing models were typically the HMC BNN, FBNN, GP-RBF and Ensemble. The variational BNNs and dropout performed less well, consistent with the results on the synthetic datasets~(see Figure~\ref{fig:toy_comparisons_batchMIG_all}). %

Using just a single model for prediction gives an incomplete picture of the query selection performance of different methods, as some methods might happen to be better matches to the NKN model or the HMC. Table~\ref{table:rank_transductive_active_learning}~(in the Appendix) shows the results of mixing and matching a wider variety of prediction and selection models. In general, we observe that regardless of which model is used for prediction, the best results are obtained when queries are selected using the most accurate models, rather than the same models used for prediction. We believe the TAL experiment indicates that high-quality posterior distributions are useful for data selection, above and beyond the benefits from making better predictions from a fixed training set. 

\textbf{Discussion. } Overall, the XLL(R) and TAL metrics present a consistent picture for evaluating PPCs. First, on both benchmarks, HMC, GP-RBF, FBNN and Ensemble are in general better than BBB, NNG and Dropout. Second, for the \texttt{boston}, \texttt{concrete} and \texttt{energy} datasets, HMC performed the best according to both metrics. Lastly, for the \texttt{wine} and \texttt{yacht} datasets, FBNN outperforms GP-RBF, reflected again by the XLL(R) and TAL performance.

%% file: tex/discussion_and_conclusion.tex
\section{Conclusion}
In this work, we developed three evaluation metrics for evaluating PPC estimators: metacorrelations, transductive active learning, and cross-normalized log likelihoods. We used synthetic data to validate that the three metrics are suitable for evaluating PPC estimators. Finally, we introduced benchmarks on real-world data to examine how accurately Bayesian models can estimate PPCs.

%% file: tex/appendix.tex
\onecolumn 
\section{Notation}
\begin{table}[h]
\centering
\caption{ The notations used in the paper. }
\label{table:notation_details}
\resizebox{1.\textwidth}{!}{
\begin{tabularx}{\textwidth}{lX}
\toprule
\textbf{Notation} & Description  \\
\midrule
$\vx$ & the input vector \\
$\{\vx_i\}_{i \in I}$ & a set of input vectors indexed by $i$ \\ 
$f$ & the underlying function \\
$\ry_{\vx}$ & the observed value of the function at a given location $\vx$ \\
$\sigma_{n}(\vx)$ & the standard variance of the observation noise at $\vx$ \\
$\sigma_n$ & the standard derivation of a homoscedastic observation noise  \\
$\sigma_{\bx}$ & standard derivation of $f(\bx)$ \\ 
$\eta$ & the prior variance for BNNs \\
\bottomrule 

\end{tabularx}
}
\end{table}

\section{The Pseudocodes}
      \begin{algorithm}[h]
        \caption{A procedure of (Transductive) Active Learning. We use {\color{red} red} and {\color{blue} blue} to show the difference between active learning and TAL. TIG and MIG can be replaced by any other acquisition functions.}
        \label{alg:active-learning}
        \begin{algorithmic}[1]
          \Require{\textit{Selection Model}: $\mathcal{M}^s$; \textit{Prediction Model}: $\mathcal{M}^p$.}
          \Require{Datasets: $\mathcal{D}_{\mathrm{tr}}=\{\mX_{\mathrm{tr}}, \vy_{\mathrm{tr}}\}$, $\mathcal{D}_{\mathrm{te}}=\{\mX_{\mathrm{te}}, \vy_{\mathrm{te}}\}$, $\mathcal{D}_{\mathrm{pl}}=\{\mX_{\mathrm{pl}}, \vy_{\mathrm{pl}}\}$.}
          \Require{Total active learning iterations: $T$; \#Queried samples per iteration: $m$.}
          \State $\mathcal{R}=\emptyset.$
          \For{$t = 1 \; \mathrm{to} \; T$}
          \State Train $\mathcal{M}^p, \mathcal{M}^s$ on $\mathcal{D}_{\mathrm{tr}}$ until convergence.
          \State Test $\mathcal{M}^p$ over $\mathcal{D}_{\mathrm{te}}$ and put the result to $\mathcal{R}$. 
          \State {\color{red} $\mathrm{InfoG} = \mathrm{TIG}(\mX_{\mathrm{pl}}, \mathcal{M}^s)$} or {\color{blue} $\mathrm{InfoG} = \mathrm{MIG}(\mX_{\mathrm{pl}}, \mX_{\mathrm{te}}, \mathcal{M}^s)$}.
          \State Sort InfoG in descending order and retrieve top $m$ samples from $\mathcal{D}_{\mathrm{pl}}$ as $\mathcal{D}_{\mathrm{qe}}$.
          \State $\mathcal{D}_{\mathrm{tr}} \leftarrow \mathcal{D}_{\mathrm{tr}} \cup \mathcal{D}_{\mathrm{qe}}$; $\mathcal{D}_{\mathrm{pl}} \leftarrow \mathcal{D}_{\mathrm{pl}} \setminus \mathcal{D}_{\mathrm{qe}}$.
          \State $t \leftarrow t + 1$.
          \EndFor
          \State \Return $\mathcal{R}, \mathcal{M}^p, \mathcal{D}_{\mathrm{tr}}$.
        \end{algorithmic}
      \end{algorithm}
      
        \begin{algorithm}[h]
        \caption{Computing XLL and XLLR.}
        \label{alg:new-metric}
        \begin{algorithmic}[1]
          \Require{Model Predictions $\{(\vmu_i, \mSigma_i)\}_{i=1}^m$; Test set $\mathcal{D}_{\text{te}}$; Batch size $b$}
          \For{$j = 1 \; \mathrm{to} \; m$} \Comment{Reference Model}
          \For{$i = 1 \; \mathrm{to} \; m$} \Comment{Normalize Predictive Marginals}
              \State $\mD^0_i=\sqrt{\diag(\mSigma_j)}/\sqrt{\diag(\mSigma_i)}$. 
              \State $\bar{\vmu}_i=\vmu_j, \bar{\mSigma}_i = \mD^0_i \mSigma_i \mD^0_i$.
          \EndFor
          \State $\mathcal{T}'=\{\}$.
          \For{$(\bx, y) \in \mathcal{D}_{\text{te}}$} \Comment{Build Test Batches}
              \State Top correlated points $\mathcal{B}_{\bx}:=\{(\bx_k, y_k)\}_{k=1}^b$; Add $\mathcal{B}_{\bx}$ to $\mathcal{T}'$.
          \EndFor
          \For{$i = 1 \; \mathrm{to} \; m$} \Comment{Compute Log Joints}
              \State $\mathrm{lld}_i^j=\frac{1}{|\mathcal{T}'|}\sum_{\mathcal{B} \in \mathcal{T}'} \log \mathcal{N}(\mathcal{B} | \bar{\vmu}_i, \bar{\mSigma}_i)$.
          \EndFor
          \State $\{\mathrm{rank}_i^j\}_{i=1}^m$ from sorting $\{\mathrm{lld}_i^j\}_{i=1}^m$ .
          \EndFor
          \State $\mathrm{lld}_i=\frac{1}{m}\sum_{j=1}^m \mathrm{lld}_i^j$, $\mathrm{rank}_i=\frac{1}{m}\sum_{j=1}^m \mathrm{rank}_i^j$ \Comment{Average over References}
          \State \Return $\{\mathrm{lld}_i\}_{i=1}^m$ and $\{\mathrm{rank}_i\}_{i=1}^m$.
        \end{algorithmic}
      \end{algorithm}

\section{Information Gains for Active Learning}\label{app:info-gain}
We introduce three types of information gains and present their analytical forms for Gaussian predictive distributions. Then, we provide a greedy approximation for computing the optimal batch corresponding to BatchMIG.

\subsection{Three Types of Information Gains}
\begin{figure}[t]
    \centering
    \includegraphics[width=0.95\textwidth]{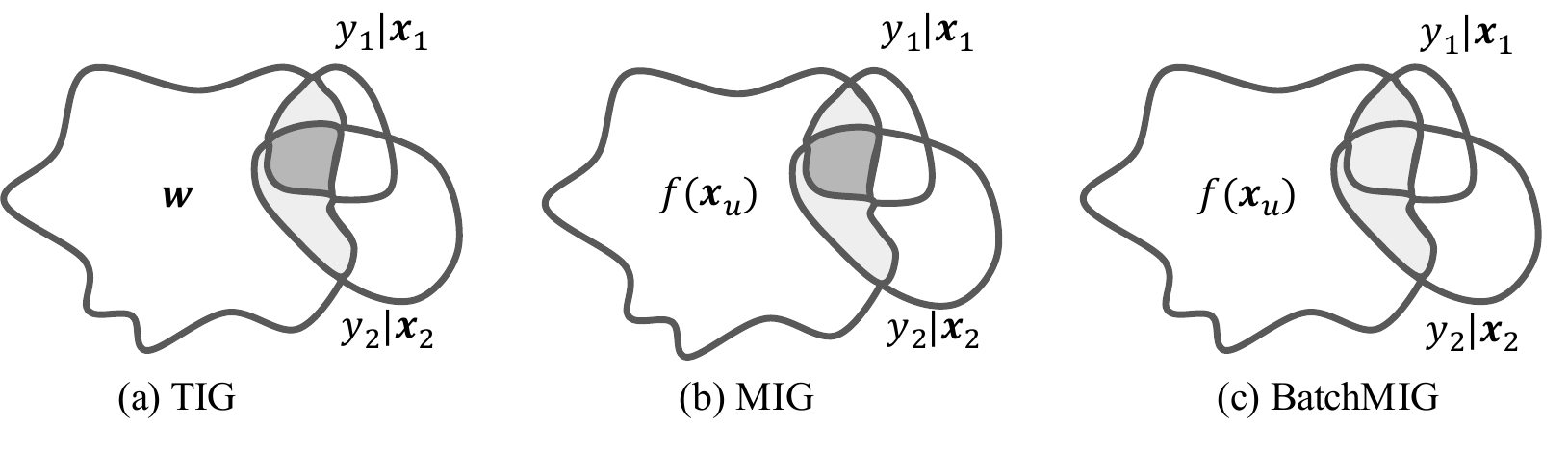}
    \caption{An illustration of how do TIG, MIG and BatchMIG compute the informativeness of two candidate points. TIG measures the mutual information between data and model, whereas MIG and BatchMIG measure that between data and test points. Dark regions represent the information gain is double-counted, \emph{i.e.,} both TIG and MIG overestimate the gain. }
    \label{fig:tal_acquisition_fig}
\end{figure}

We firstly specify the analytic expressions for computing the information gain acquisition functions:

\textbf{Total Information Gain (TIG)}, measures the mutual information between the queried point $\bx$ and the model parameters $\bw$, 
    \begin{align}
    \mathrm{TIG}(\vx) &\coloneqq \mathbb{I}(y_{\bx}; \bw |\dataset_{\mathrm{tr}})
     \overset{\text{Gaussian predictive dist}}{=} \frac{1}{2}\log\left(1 + \sigma_\vx^2/\sigma_n(\bx)^2 \right) ,
\end{align}

\textbf{Marginal Information Gain (MIG)}, measures the mutual information between the queried point $\bx$ and a point $\bx_u$ of interest,
\begin{align}
    &\mathrm{MIG}(\vx;\vx_{u}) \coloneqq \mathbb{I}(y_{\bx}; f(\bx_u) |\dataset_{\mathrm{tr}})
    \overset{\text{Gaussian predictive dist}}{=}  -\frac{1}{2}\log\left(1 - \frac{\mathrm{Cov}(\vx, \vx_u)^2}{\sigma^2_{\vx_u}(\sigma_{\vx}^2 + \sigma_n(\vx)^2 }\right),
\end{align}
\textbf{Batched Marginal Information Gain (BatchMIG)}, measures the mutual information between a batch of queried points $\bx_{1:q}$ and the point $\bx_u$ of interest,
\begin{align}
\label{eq:BMIG}	
    &\textrm{BatchMIG}(\vx_{1:q}; \vx_u) 
    \coloneqq \mathbb{I}(y_{\bx_{1:q}}; f(\bx_u) |\dataset_{\mathrm{tr}}) \notag \\
    &\overset{\text{Gaussian predictive dist}}{=}  -\frac{1}{2}\log\left(1 - \frac{\mathrm{Cov}\left(\vx_{1:q}, \vx_u\right)^\top\left(\mathrm{Cov}\left(\vx_{1:q}, \vx_{1:q}\right) + \bsigma^2_n(\vx_{1:q})  \right)^{-1}\mathrm{Cov}\left(\vx_{1:q}, \vx_u\right)}{\sigma_{\vx_u}^2}\right),
\end{align}

Again for MIG and BatchMIG, assuming that we are interested at a set of points $\{\bx_u^i\}_{i=1}^I$, as recommended in \citet{mackay1992information}, we adopt the mean marginal information gains: $\frac{1}{I}\sum_{i=1}^I  \mathrm{MIG}(\vx;\vx_{u}^i)$ and $\frac{1}{I}\sum_{i=1}^I  \mathrm{BatchMIG}(\vx;\vx_{u}^i)$.

\subsection{A Greedy Approximation of the Optimal Batch}
In practice we will usually query a batch of points at each iteration for efficiency. For TIG and MIG, selecting a batch corresponds to selecting the points with highest information gains, correspondingly. For BatchMIG, although extending the information gain acquisition functions from the single-point scenario to the batch scenario is straightforward, solving for the optimal batch requires a combinatorial explosion of computations. Following \citep{kirsch2019batchbald}, we adopt a greedy approximation of the optimal batch, which is specified in Alg~\ref{alg:batch-bald}.

\begin{algorithm}[h]
        \caption{BatchBald \citep{kirsch2019batchbald}: a greedy approximation of the optimal batch.}
        \label{alg:batch-bald}
        \begin{algorithmic}[1]
          \Require{Model $\mathcal{M}$, Points of interest $\mathcal{I}$, Query Batch Size $q$}
          \Require{The information gain acquisition function $\mathrm{IG}$.}
          \State $A \leftarrow \emptyset$
          \For{$i = 1 \; \mathrm{to} \; q$}
          \State $\bx^{\star} \in \arg \max_{\bx} \mathrm{IG}(\{\bx\}\cup A, \mathcal{M}, \mathcal{I})$
          \State $A = A \cup \{\bx^{\star}\}$
          \EndFor
          \State \Return $A$.
        \end{algorithmic}
      \end{algorithm}

\section{Experimental Details}

\subsection{Hyperparameters}
We use the standard regression task for tuning hyperparameters with respect to each method and each dataset. Specifically, we split the dataset into train ($60\%$), valid ($20\%$) and test ($20\%$). Across 10 different runs, we use the same validation set but split train/test randomly. Finally the averaged validation log likelihood will be used for selecting the hyperparameters. A list of details about hyperparameters is shown in Table~\ref{table:hp_tune_details}.

With the tuned hyperparameters, we conduct transductive active learning and compute the XN-LLDR metrics. To avoid that the test set being used for tuning hyper-parameters, we make sure the randomly selected test set is disjoint with the validation set for hyperparameter tuning.

\begin{table}[h]
\centering
\caption{\small The hyperparameters for each method  }
\label{table:hp_tune_details}
\resizebox{1.0\textwidth}{!}{
\begin{tabularx}{\textwidth}{lX}
\toprule
\textbf{Methods} & Hyperparameters to tune  \\
\midrule
BBB &  lr: $[0.001, 0.003, 0.01]$, hidden units: $[50, 400]$, \#eval\_cov\_samples: $[100, 700, 5000]$ \\
NNG &  lr: $[0.001, 0.003, 0.01]$, hidden units: $[50, 400]$ \\
HMC &  lr: $[0.001, 0.003, 0.01]$, hidden units: $[50, 400]$ \\
FBNN & lr: $[0.001, 0.003, 0.01]$, number of random measurement points $[5, 20, 100]$, hidden units: $[50, 400]$ \\
Dropout & lr: $[0.001, 0.003, 0.01]$, hidden units: $[50, 400]$, Dropout Rate: $[0.0025, 0.01, 0.05]$, \\ & Observation variance: $[0.005, 0.025, 0.125]$\\
Ensemble & lr: $[0.001, 0.003, 0.01]$, hidden units: $[50, 400]$ \\
\midrule
\textbf{Methods} & Other Settings \\
\midrule
(SV)GP &  Optimizer=Adam, lr=0.003, epochs=10,000, batch\_size=$\min(5,000, \#\textrm{training data})$, lenghth\_scale are initialized with k-means on training data, ARD=True, min\_obsvar=1e-5~(except for \texttt{Wine} dataset, we use min\_obsvar = 1e-8); For large datasets, we adopt SVGP with 1,000 inducing points; For (SV)GP-NKN, we adopt the same NKN as in~\citet{sun2018differentiable} and epochs=5,000.\\
BBB &  Optimizer=Adam, epochs=10,000, batch\_size=100, \#training\_particles=10, \#test\_particles=5,000. \\
NNG &   Optimizer=NG-KFAC(damping=1e-5, ema\_cov\_decay=0.999), epochs=10,000, lr decay by a factor 0.1 every 5000 epochs, \#training\_particles=10, \#test\_particles=5,000, \#eval\_cov\_samples=5000. \\
HMC &  \#chains = 10, burnin=5,000 for small datasets and 15,000 for larger ones, step\_size starts at 0.01 but is adapted according to the acceptance rate, \#leap\_frog\_steps=5; We select one particle every 100 samples after burnin untill we collected 100 samples in each chain, which results at 1,000 samples for testing and computing the covariance. We use Adam Optimizer for optimizing the prior hyperparameters $\eta, \xi$ every 10 HMC steps.\\
FBNN &  Optimizer=Adam, epochs=10,000, batch\_size=\#training data for small datasets and $900$ for larger datasets in order to match the computation complexity of SVGP. The network has 400 hidden units with cosine activations.\\
Dropout & Optimizer=Adam, epochs=10,000, batch\_size=100. We use $5,000$ samples for test and computing the covariance. L2 regularization with $10^{-4} * (1 - \mathrm{dropout\_rate}) / (2. * N * \xi)$.\\
Ensemble & Optimizer=Adam, epochs=10,000, batch\_size=100, \#networks=100. \\
\bottomrule

\end{tabularx}
}
\vspace{-0.4cm}
\end{table}

\newpage
\section{Additional Results}
We present here the additional results, including (1) Log Joints versus Log Marginals; (2) Average Rank in TAL; (3) Average Log Joint Likelihood on UCI datasets; (4) RMSE performance of TAL using different Acquisition functions; (5) Comparisons between Different Data Acquisition Functions; (6) TAL Results of Different Models on Synthetic Dataset.

\subsection{Log Joints versus Log Marginals}
\begin{figure}[h]
    \centering
    \includegraphics[width=1.0\columnwidth]{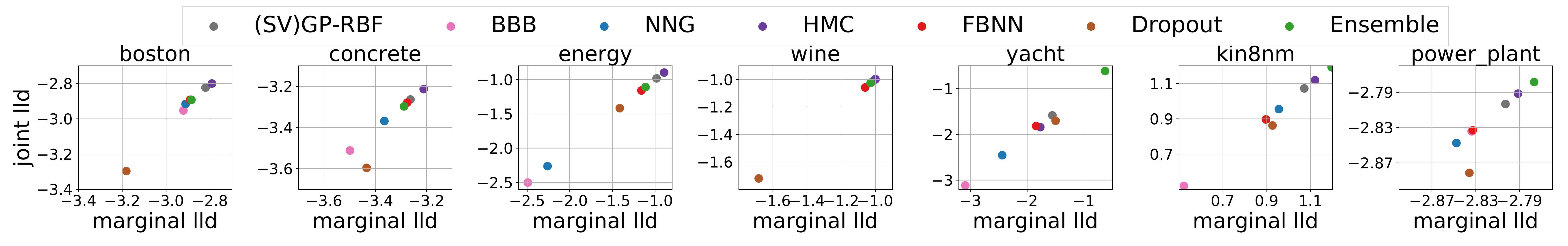}
    \caption{Scatters of log joint likelihoods versus log marginal likelihoods where each point represents one method. The log joints are computed over random batches with 5 points.}
    \label{fig:app:marginal_joint}
\end{figure}

We visualize the scatter plot of the joint log-likelihoods and the marginal log-likelihoods in Figure~\ref{fig:app:marginal_joint}. We observe that the joint log-likelihood is positively correlated with the marginal log-likelihood.

\subsection{Average Rank in TAL (Table~\ref{table:rank_transductive_active_learning})}

\begin{table*}[h]
\centering
\caption{\small Average rank of each method's LLD and RMSE on TAL at the last iteration with different prediction models. We use {\color{red}red} to highlight the best ones, and {\color{blue}blue}  for the worst ones.}
\label{table:rank_transductive_active_learning}
\resizebox{0.95\textwidth}{!}{
\begin{tabular}{l cccc cccc}
\toprule
& \textbf{Prediction Model/Selection Model} &(SV)GP-RBF & BBB & NNG & HMC & FBNN & Dropout & Ensemble
\\

\midrule
\multirow{5}{*}{RMSE}&
Oracle
& 
2.4 & 
{\color{blue}4.3} & 
3.8 & 
{\color{red}2.0} & 
2.4 & 
3.0 & 
3.0 
 
\\
&Dropout
&
2.8 & 
{\color{blue}4.4} & 
4.3 & 
{\color{red}1.4} & 
2.4 & 
2.5 & 
3.2  
\\
&(SV)GP-RBF
& 
2.4 & 
{\color{blue}4.3} & 
3.7 & 
{\color{red}1.6} & 
2.0 & 
3.2 & 
3.8 
\\
&NNG &
2.7 & 
{\color{blue}3.9} & 
3.7 & 
{\color{red}1.9} & 
2.1 & 
3.2 & 
3.5 
\\
&HMC &
2.4 & 
{\color{blue}4.4} & 
3.4 & 
2.5 & 
{\color{red}1.8} & 
3.4 & 
3.1 
\\
\midrule
&
\textbf{Average Rank}
& 2.5
& {\color{blue}4.3}
& 3.8
& {\color{red}1.9}
& 2.1
& 3.1
& 3.3
\\
\midrule

\multirow{5}{*}{LLD}&
Oracle
&
2.1 & 
{\color{blue}4.5} & 
4.0 & 
{\color{red}1.8} & 
2.2 & 
4.0 & 
2.5 

\\
&Dropout
&
2.8 & 
{\color{blue}4.5} & 
4.1 & 
{\color{red}1.8} & 
2.6 & 
2.6 & 
2.6 

\\
&(SV)GP-RBF
&
2.5 & 
{\color{blue}4.2} & 
3.8 & 
{\color{red}1.7} & 
2.1 & 
3.3 & 
3.3 
\\
&NNG &
2.7 & 
{\color{blue}3.9} & 
3.8 & 
2.1 & 
{\color{red}2.0} & 
3.3 & 
3.3 

\\
&HMC &
2.8 & 
{\color{blue}4.5} & 
3.2 & 
2.5 & 
{\color{red}2.0} & 
3.9 & 
2.2 

\\
\midrule
&
\textbf{Average Rank}
& 2.6
& {\color{blue}4.3} 
& 3.8
& {\color{red}2.0}
& 2.2
& 3.4
& 2.8
\\
\bottomrule

\end{tabular}
}
\vspace{-0.2cm}
\end{table*}

Table~\ref{table:rank_transductive_active_learning} shows the results of mixing and matching a wider variety of training and selection models. In general, we observe that regardless of which model is used for training, the best results are obtained when queries are selected using the most accurate models, rather than the same models used for training. We believe this experiment directly indicates that high-quality posterior predictive distributions are useful for data selection, above and beyond the benefits from making better predictions from a fixed training set. 

\newpage

\subsection{Average XLLs on UCI datasets~(Table~\ref{table:pearson_corr})}\label{app:XN-LLD}

\begin{table*}[h]
\centering
\caption{The average XLL for each model on UCI datasets. }
\label{table:pearson_corr}
\resizebox{\textwidth}{!}{
\begin{tabular}{lccccccc}
\toprule
\textbf{Dataset}/\textbf{Method} 
& (SV)GP-RBF & BBB & NNG & HMC & FBNN & Dropout & Ensemble
\\

\midrule 
\texttt{Boston}
&-3.217 (0.134)
&-3.316 (0.156)
&-3.202 (0.133)
&-3.177 (0.133)
&-3.237 (0.138)
&-3.456 (0.160)
&-3.202 (0.139)
\\
\texttt{Concrete}
&-3.342 (0.015)
&-3.394 (0.018)
&-3.351 (0.016)
&-3.336 (0.015)
&-3.344 (0.015)
&-3.615 (0.029)
&-3.340 (0.015)
\\

 \texttt{Energy}
&-1.382 (0.065)
&-1.430 (0.068)
&-1.437 (0.068)
&-1.378 (0.064)
&-1.384 (0.064)
&-1.434 (0.067)
&-1.386 (0.065)
\\
 \texttt{Wine}
&-1.215 (0.032)
&-1.266 (0.038)
&-1.228 (0.034)
&-1.224 (0.034)
&-1.222 (0.035)
&-1.306 (0.042)
&-1.226 (0.034)
\\
 \texttt{Yacht}
&-2.062 (0.115)
&-2.112 (0.108)
&-2.074 (0.102)
&-2.126 (0.118)
&-2.011 (0.103)
&-2.674 (0.166)
&-1.998 (0.102)
\\

 \texttt{Kin8nm}
&0.902 (0.031)
&0.892 (0.031)
&0.890 (0.032)
&0.902 (0.031)
&0.901 (0.031)
&0.796 (0.032)
&0.897 (0.031)
\\

\texttt{Naval}
&6.853 (0.172)
&6.795 (0.176)
&6.811 (0.175)
&6.882 (0.166)
&6.870 (0.171)
&6.971 (0.163)
&6.920 (0.173)
\\

 \texttt{Power\_plant}
&-2.793 (0.015)
&-2.812 (0.018)
&-2.821 (0.019)
&-2.801 (0.017)
&-2.806 (0.017)
&-2.828 (0.015)
&-2.796 (0.016)
\\
\bottomrule

\end{tabular}
}
\end{table*}

\subsection{RMSE performance of TAL using different Acquisition functions. }
\begin{figure*}[h]
    \centering
    \includegraphics[width=0.49\textwidth]{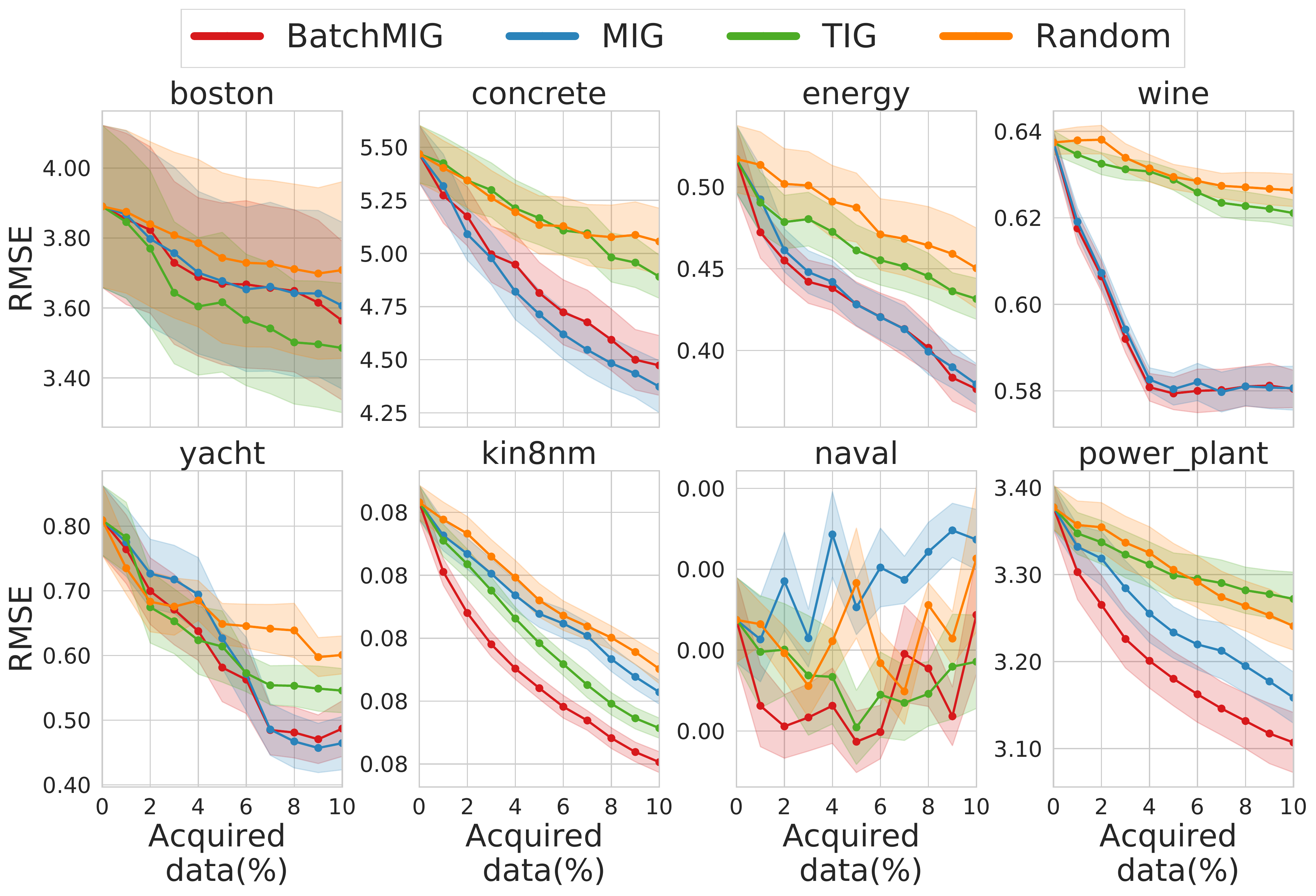}
    \hfill
    \includegraphics[width=0.49\textwidth]{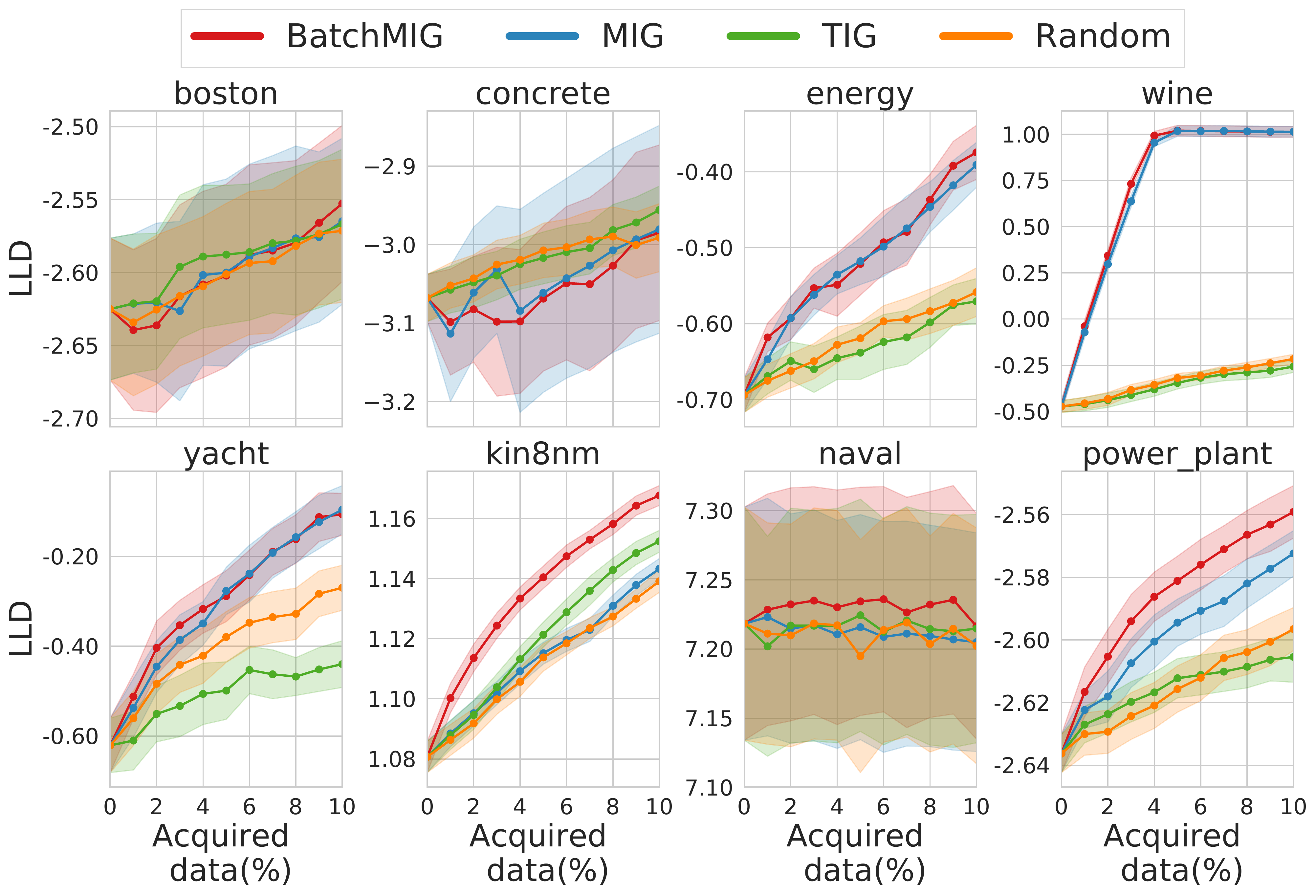}
    \caption{RMSE and LLD performance of TAL with the pre-trained NKN kernel~(Oracle).}
    \label{fig:app:rmse_active_learning_Oracle}
\end{figure*}

In addition to the LLD performance, we also present the RMSE performance of TAL using BatchMIG, MIG, TIG and random selection with the `Oracle' model in Figure~\ref{fig:app:rmse_active_learning_Oracle}~(right part).

\newpage

\subsection{Comparisons between Different Data Acquisition Functions}
\begin{figure*}[h]
\centering
\includegraphics[width=0.6\textwidth]{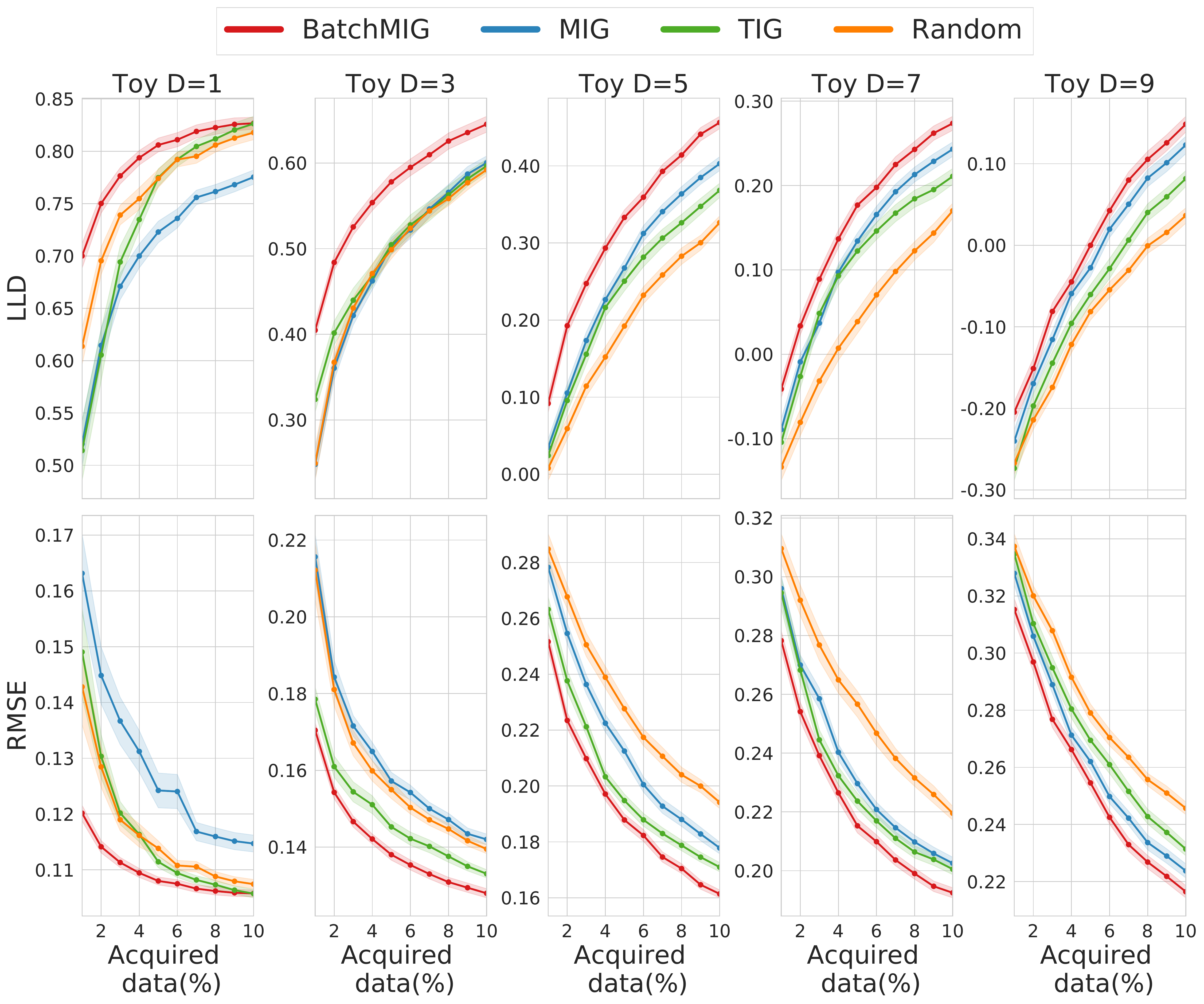}
       \caption{Comparisons between different acquisition functions with Oracle model.}	
       \label{fig:toy_comparisons_acquisition_oracle_all}
\end{figure*}

\begin{figure}[h]
    \centering
     \includegraphics[width=0.6\textwidth]{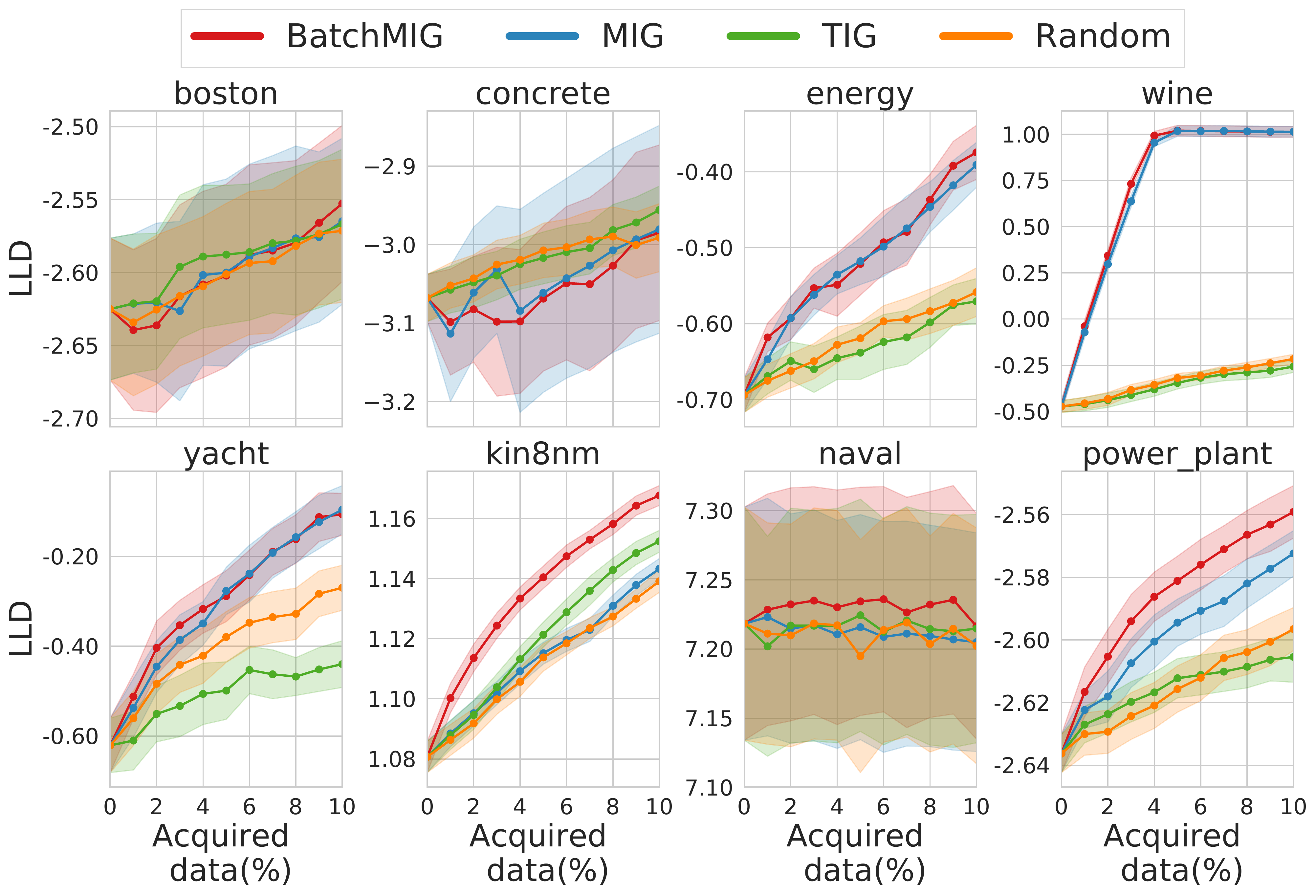}
    \caption{Right: Comparing TAL criteria on UCI datasets using the Oracle (NKN) model.\label{app:fig:compare_criterion_uci} }
\end{figure}

We present here the results using different data acquisition functions on synthetic datasets and on UCI datasets. The results can be found in Figure~\ref{fig:toy_comparisons_acquisition_oracle_all} and Figure~\ref{app:fig:compare_criterion_uci}, where we can observe that TAL acquisition functions consistently outperform other criterions.

\newpage
\subsection{TAL Results of Different Models on Synthetic Datasets}
\begin{figure*}[h]
\centering
\vspace{-0.3cm}
\includegraphics[width=\textwidth]{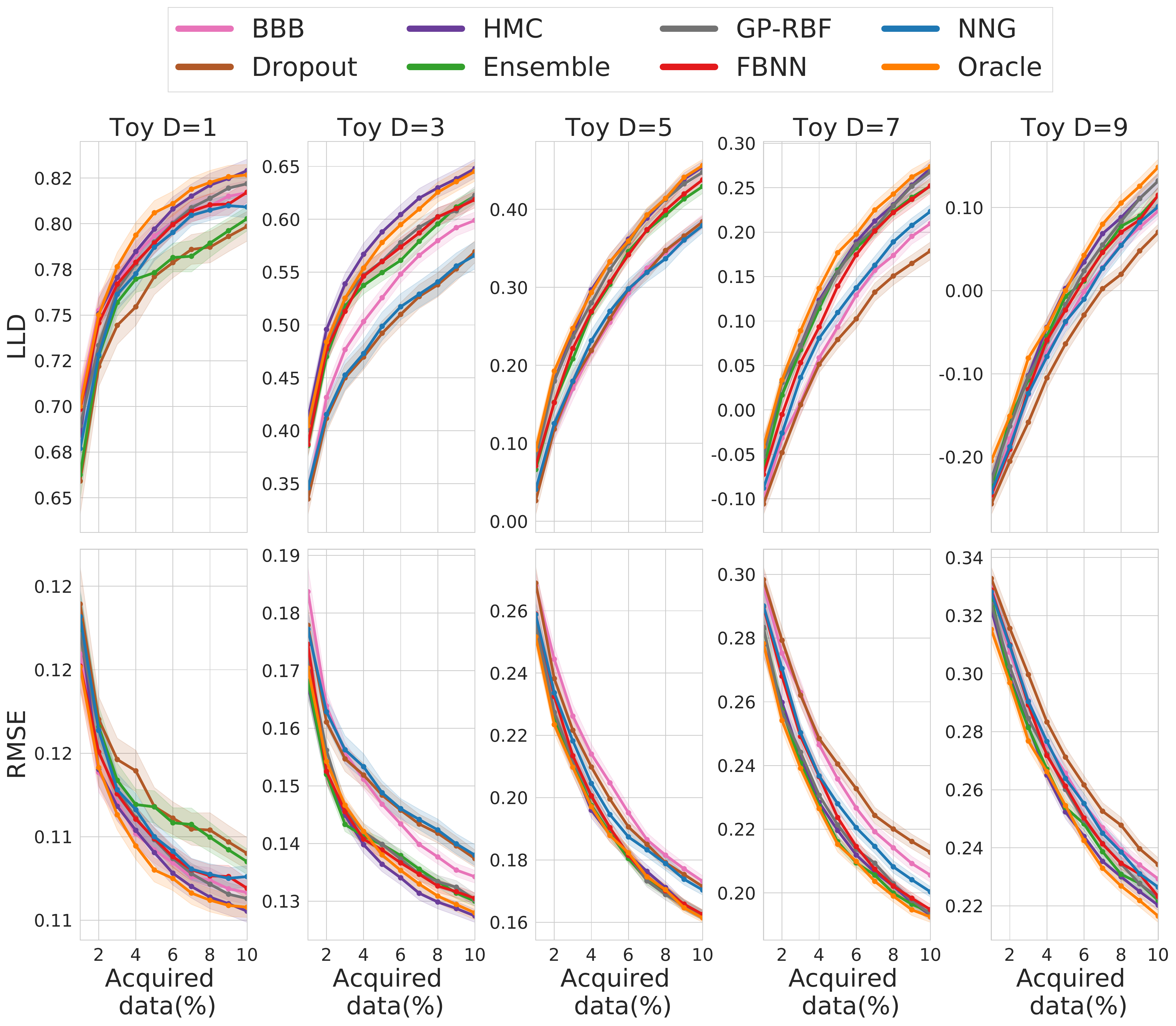}
       \caption{BatchMIG on toy datasets, with fixed observation variance.}	
       \label{fig:toy_comparisons_batchMIG_all}
       \vspace{-0.8cm}
\end{figure*}

\begin{figure*}[h]
\centering
\includegraphics[width=\textwidth]{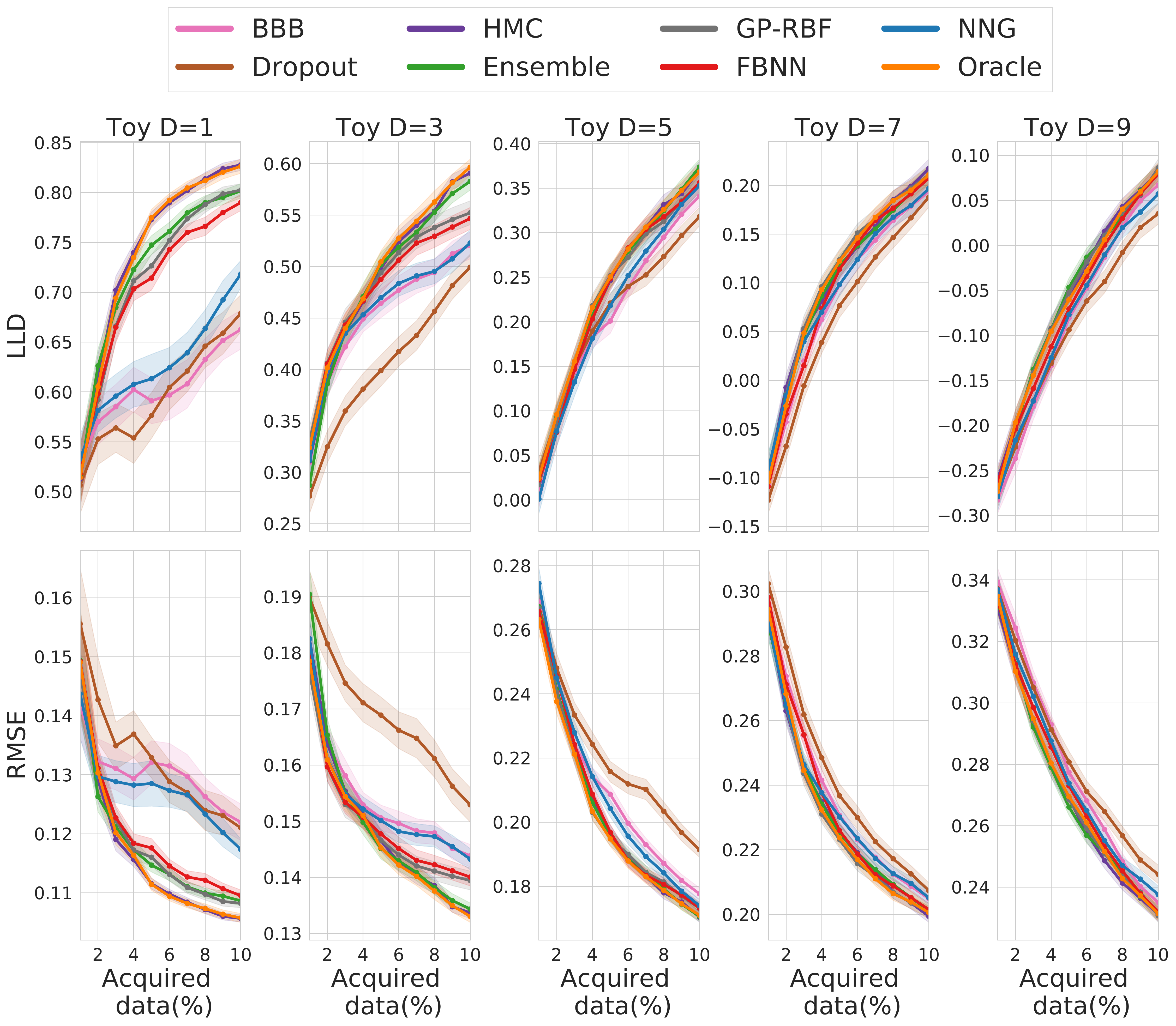}
       \caption{TIG on toy datasets, with fixed observation variance.}	
       \label{fig:toy_comparisons_TIG_all}
       \vspace{-0.4cm}
\end{figure*}

To evaluate how each models perform on TAL, we compare them with BatchMIG and TIG on the synthetic datasets. The results are presented in Figure~\ref{fig:toy_comparisons_batchMIG_all} and Figure~\ref{fig:toy_comparisons_TIG_all} respectively.

\newpage

\newpage

\section{A Theoretical Connection between Log Likelihoods and Predictive Correlations}

To understand why XLL directly reflects the accuracy of the correlations, consider the following distributions: 
\begin{align} \label{app:eq:def_distributions}
	&p(\vy|\mX) = \mathcal{N}(\vy|\vmu_\text{gen}, \diag(\bsigma_\text{gen})\mC_\text{gen}\diag(\bsigma_\text{gen})), \notag \\
	&q(\vy|\mX) = \mathcal{N}(\vy|\vmu_{\text{ref}},\diag(\bsigma_{\text{ref}})\mC\diag(\bsigma_{\text{ref}})), \notag \\
	&p_m(\vy|\mX) = \mathcal{N}(\vy|\vmu_\text{gen},\diag(\bsigma_\text{gen})), \notag \\
	&q_m(\vy|\mX) = \mathcal{N}(\vy|\vmu_{\text{ref}},\diag(\bsigma_{\text{ref}})),  \notag \\
	&p_c(\vy|\mX) = \mathcal{N}(\vy|\vzero, \mC_{\text{gen}}), \notag \\
	&q_c(\vy|\mX) = \mathcal{N}(\vy|\vzero, \mC),
\end{align}
where $p(\vy|\mX)$ is the data generating distribution, and $\vmu_{\text{gen}}$, $\bsigma_{\text{gen}}^2$ and $\mC_{\text{gen}}$ are the ground-truth mean, variance and correlations respectively. %
Observe that $-\KL{p}{q}$ is the quantity that XLL is approximating using samples (up to a constant), while $\KL{p_c}{q_c}$ is a measure of dissimilarity between the correlation matrices or the LogDet divergence between two positive semidefinite matrices $\mC_{\text{gen}}$ and $\mC$. We now show that, if the reference marginals (i.e., $\bmu_{\text{ref}}, \bsigma_{\text{ref}}$) are close to the ground truth marginals, then $\KL{p}{q}$ approximately equals $\KL{p_c}{q_c}$. Hence, XLL can be seen as a measure of the accuracy of the predictive correlations.

\begin{theorem}\label{thm:corr-est} Let the predictive distributions be defined above, and let $b$ be the number of points for evaluation, $\lambda$ denote the smallest eigenvalue of $\bC$ and $\xi = \KL{p_m}{q_m}$. If $\xi \ll 1$, then we have:
  \begin{align}
     \abs{\KL{p}{q} - \KL{p_c}{q_c}} = \mathcal{O}\left(\frac{b^{3/2}}{\lambda}\sqrt{\xi} \right).
\end{align}

\end{theorem}

\begin{remark} Because the expected joint log-likelihood $\expect_{p(\vy|\mX)}\log q(\vy|\mX)= \expect_{p(\vy|\mX)}\log p(\vy|\mX) - \KL{p}{q}$, this theorem illustrates that, for nearly-optimal reference marginals, the expected joint log-likelihood reflects the quality of the predictive correlations. This validates the reliability of XLL. %
\end{remark}
\begin{remark}
In practice, the predictive covariance is $\bSigma + \sigma_{n}^2 \bI$, where $\sigma_{n}^2$ is the variance of the modeled observation noise and $\bSigma$ is the predictive covariance for the underlying function. In general, $\sigma_{n}^2$ and the predictive variances in $\bSigma$ are in the same order of magnitude. Therefore, the smallest eigenvalue $\lambda$ of the correlation matrix $\bC$ is not much smaller than $1$. Furthermore, $b$ is small because we evaluate XLL and XLLR over mini-batches ($b=5$ in our experiments).
\end{remark}

As suggested by the theorem, the ideal reference model would be the oracle, i.e.~the true data generating distribution. However, in practice we only have access to models which imperfectly match the distribution. Fortunately, we found that the relative order of XLL values do not appear to be overly sensitive to the choice of reference model.
Therefore, to further avoid favoring any particular model as the reference, we propose to iterate through every candidate model to act as the reference model once. Then, for each candidate model, we average its XLL or XLLR across all reference models. Empirically, we found that XLL and XLLR align well with the corresponding performance in TAL benchmarks as well as the oracle-based meta-correlations. In below, we provide the proof of Theorem~\ref{thm:corr-est}:

\newpage
\begin{proof}

We first define: 
\begin{gather}
	\vd := \frac{\vmu_{\text{gen}} - \vmu_{\text{ref}}}{\bsigma_{\text{ref}}},
	\quad \vr:=\frac{\bsigma_{\text{gen}}}{\bsigma_{\text{ref}}}, %
\end{gather}
and let $\bone \in \Reals^{b \times b}$ be the all-ones matrix and $\bI$ be the identity matrix, then we have:
\begin{align}
    2\KL{p}{q} =& \log \frac{\abs{\mathrm{diag}(\bsigma_{\text{gen}})\mC\mathrm{diag}(\bsigma_{\text{gen}})}}{\abs{\mathrm{diag}(\bsigma_{\text{ref}})\mC_{\text{gen}}\mathrm{diag}(\bsigma_{\text{ref}})}} - b + \trace\left(\mC^{-1} \mathrm{diag}(\vr)\mC_{\text{gen}}\mathrm{diag}(\vr) \right) + \vd^{\top}\mC^{-1} \vd \notag \\
    =& \sum_{i=1}^b \log\frac{\bsigma_{\text{ref}, i}^2}{\bsigma_{\text{gen},i}^2} + \log \frac{\abs{\mC}}{\abs{\mC_{\text{gen}}}} - b + \vr^{\top} \left(\mC^{-1} \circ \mC_{\text{gen}} \right)  \vr + \vd^{\top}\mC^{-1} \vd \notag \\
    =& \underbrace{\log \frac{\abs{\mC}}{\abs{\mC_{\text{gen}}}} - b + \trace\left(\left(\mC^{-1} \circ \mC_{\text{gen}} \right) \bone   \right)}_{2\KL{p_c}{q_c}}  - \underbrace{\sum_{i=1}^b \log \vr_i^2 + \vd^{\top} \vd + \vr^{\top} \vr - b}_{-2\KL{p_m}{q_m}} \notag \\
    &+ \underbrace{\trace\left(\left(\mC^{-1} \circ \mC_{\text{gen}} \right)\left( \vr\vr^{\top} - \bone \right)  \right)}_{\circled{1}}  + \underbrace{\trace \left(\left(\mC^{-1}-\bI\right) \vd\vd^{\top} \right)}_{\circled{2}} + \underbrace{\left(b - \vr^{\top} \vr\right)}_{\circled{3}}.
\end{align}
Therefore, we have
\begin{align}
    2 \abs{\KL{p_c}{q_c} - \KL{p}{q}} \leq  &\underbrace{2\KL{p_m}{q_m}}_{2\xi} + \abs{ \circled{1}} + \abs{\circled{2}} + \abs{\circled{3}},
\end{align}
Given that the marginal KL divergence is upper bounded by,
\begin{align}
    2\KL{p_m}{q_m} = - \sum_{i=1}^b \log \vr_i^2 + \vd^{\top} \vd + \vr^{\top} \vr - b \leq 2\xi,
\end{align}
and since $\forall x, x - 1 - \log x \geq 0$, we have
\begin{align}
	0 \leq - \sum_{i=1}^b \log \vr_i^2  + \vr^{\top} \vr - b \leq 2\xi.
\end{align}
Then $\forall i$, $\vr_i^2 - \log  \vr_i^2 - 1 \leq 2 \xi $, which means $\vr_i = 1 + \mathcal{O}(\sqrt{\xi})$. As a result, we have the following bounds,
\begin{align}
	&\|\vd\vd^{\top}\|_F = \vd^{\top} \vd \leq 2 \xi, \\
    &\abs{b - \vr^{\top} \vr } = \mathcal{O}(b \sqrt{\xi}), \\
    &\| \vr\vr^{\top} - \bone \|_F = \mathcal{O}(b \sqrt{\xi}).
\end{align}

We further let $\lambda:=\lambda_{min}(\mC)$ be the smallest eigenvalue of $\mC$. Then, we have $\|\mC^{-1}\|_2 = \frac{1}{\lambda}$. Because $\mC_{\text{gen}}$ is a correlation matrix, $\|\mC_{\text{gen}}\|_{\infty}=1$. Because 
\begin{align}
	\left(\trace\left(\mA^{\top}\mB\right)\right)^2 \leq \trace\left(\mA^{\top}\mA\right)\trace\left(\mB^{\top}\mB\right)=\|\mA\|_F^2\|\mB\|_F^2,
\end{align}
which gives us the upper bound of {\small$\abs{\circled{1}}$}:
\begin{align}
    \abs{\circled{1}} =& \abs{ \trace\left(\left(\mC^{-1} \circ \mC_{\text{gen}} \right)\left( \vr\vr^{\top} - \bone \right)  \right)} \notag \\
     \leq& \| \mC^{-1} \circ \mC_{\text{gen}} \|_F \| \vr\vr^{\top} - \bone \|_F \notag \\
     \leq& \|\mC^{-1}\|_F \| \vr\vr^{\top} - \bone \|_F \notag \\
     \leq& \frac{\sqrt{b}}{\lambda}\| \vr\vr^{\top} - \bone \|_F, \notag \\
     =& \frac{\sqrt{b}}{\lambda} \mathcal{O}(b\sqrt{\xi}).
\end{align}
Similarly, we can further bound {\small $\abs{\circled{2}}$} by:
\begin{align}
    \abs{\circled{2}} =& \abs{\trace \left(\left(\mC^{-1}-\bI\right) \vd\vd^{\top} \right) } \notag \\ 
    \leq&  \|\mC^{-1}-\bI\|_F \|\vd\vd^{\top}\|_F \notag \\
     \leq& \sqrt{2b + 2 \|\mC^{-1}\|_F^2} \|\vd\vd^{\top}\|_F \notag \\
     \leq& \sqrt{2b + \frac{2b}{\lambda^2}} \|\vd\vd^{\top}\|_F \notag \\
     \leq& \sqrt{2b + \frac{2b}{\lambda^2}} 2\xi.
\end{align}

Lastly, we can bound:
\begin{align}
	\abs{\circled{3}} = \abs{b - \vr^{\top} \vr} = \mathcal{O}(b \sqrt{\xi}).
\end{align} 

Overall, since $\xi \ll 1$, we have
\begin{align}
     2 \abs{\KL{p_c}{q_c} - \KL{p}{q}} \leq& 2\KL{p_m}{q_m} +  \frac{\sqrt{b}}{\lambda} \mathcal{O}(b \sqrt{\xi}) + \sqrt{2 + \frac{2b}{\lambda^2}} 2 \xi + \mathcal{O}(b \sqrt{\xi})  \notag \\
     \leq& 2\xi +  \frac{\sqrt{b}}{\lambda} \mathcal{O}(b \sqrt{\xi}) + \sqrt{2 + \frac{2b}{\lambda^2}} 2 \xi + \mathcal{O}(b \sqrt{\xi})  \notag \\
     =& \mathcal{O}\left(\frac{b^{3/2}}{\lambda}\sqrt{\xi} \right).
\end{align}

\end{proof}

%% file: sample_paper.bbl
\begin{thebibliography}{36}
\providecommand{\natexlab}[1]{#1}
\providecommand{\url}[1]{\texttt{#1}}
\expandafter\ifx\csname urlstyle\endcsname\relax
  \providecommand{\doi}[1]{doi: #1}\else
  \providecommand{\doi}{doi: \begingroup \urlstyle{rm}\Url}\fi

\bibitem[Beluch et~al.(2018)Beluch, Genewein, N{\"u}rnberger, and
  K{\"o}hler]{beluch2018power}
Beluch, W.~H., Genewein, T., N{\"u}rnberger, A., and K{\"o}hler, J.~M.
\newblock The power of ensembles for active learning in image classification.
\newblock In \emph{Proceedings of the IEEE Conference on Computer Vision and
  Pattern Recognition}, pp.\  9368--9377, 2018.

\bibitem[Blundell et~al.(2015)Blundell, Cornebise, Kavukcuoglu, and
  Wierstra]{blundell2015weight}
Blundell, C., Cornebise, J., Kavukcuoglu, K., and Wierstra, D.
\newblock Weight uncertainty in neural network.
\newblock In \emph{International Conference on Machine Learning}, pp.\
  1613--1622, 2015.

\bibitem[Cobb et~al.(2018)Cobb, Roberts, and Gal]{cobb2018loss}
Cobb, A.~D., Roberts, S.~J., and Gal, Y.
\newblock Loss-calibrated approximate inference in bayesian neural networks.
\newblock \emph{arXiv preprint arXiv:1805.03901}, 2018.

\bibitem[Cohn et~al.(1996)Cohn, Ghahramani, and Jordan]{cohn1996active}
Cohn, D.~A., Ghahramani, Z., and Jordan, M.~I.
\newblock Active learning with statistical models.
\newblock \emph{Journal of artificial intelligence research}, 4:\penalty0
  129--145, 1996.

\bibitem[Delyon et~al.(1999)Delyon, Lavielle, and
  Moulines]{delyon1999convergence}
Delyon, B., Lavielle, M., and Moulines, E.
\newblock Convergence of a stochastic approximation version of the em
  algorithm.
\newblock \emph{Annals of statistics}, pp.\  94--128, 1999.

\bibitem[Filos et~al.(2019)Filos, Farquhar, Gomez, Rudner, Kenton, Smith,
  Alizadeh, de~Kroon, and Gal]{oatml2019bdlb}
Filos, A., Farquhar, S., Gomez, A.~N., Rudner, T. G.~J., Kenton, Z., Smith, L.,
  Alizadeh, M., de~Kroon, A., and Gal, Y.
\newblock Benchmarking bayesian deep learning with diabetic retinopathy
  diagnosis.
\newblock \url{https://github.com/OATML/bdl-benchmarks}, 2019.

\bibitem[Frazier et~al.(2009)Frazier, Powell, and
  Dayanik]{frazier2009knowledge}
Frazier, P., Powell, W., and Dayanik, S.
\newblock The knowledge-gradient policy for correlated normal beliefs.
\newblock \emph{INFORMS journal on Computing}, 21\penalty0 (4):\penalty0
  599--613, 2009.

\bibitem[Gal \& Ghahramani(2016)Gal and Ghahramani]{gal2016dropout}
Gal, Y. and Ghahramani, Z.
\newblock Dropout as a {B}ayesian approximation: Representing model uncertainty
  in deep learning.
\newblock In \emph{International Conference on Machine Learning}, pp.\
  1050--1059, 2016.

\bibitem[Gal et~al.(2017)Gal, Islam, and Ghahramani]{gal2017deep}
Gal, Y., Islam, R., and Ghahramani, Z.
\newblock Deep bayesian active learning with image data.
\newblock In \emph{International Conference on Machine Learning}, pp.\
  1183--1192, 2017.

\bibitem[Guo et~al.(2017)Guo, Pleiss, Sun, and Weinberger]{guo2017calibration}
Guo, C., Pleiss, G., Sun, Y., and Weinberger, K.~Q.
\newblock On calibration of modern neural networks.
\newblock In \emph{International Conference on Machine Learning}, pp.\
  1321--1330, 2017.

\bibitem[Hennig \& Schuler(2012)Hennig and Schuler]{hennig2012entropy}
Hennig, P. and Schuler, C.~J.
\newblock Entropy search for information-efficient global optimization.
\newblock \emph{Journal of Machine Learning Research}, 13\penalty0
  (Jun):\penalty0 1809--1837, 2012.

\bibitem[Hensman et~al.(2013)Hensman, Fusi, and Lawrence]{hensman2013gaussian}
Hensman, J., Fusi, N., and Lawrence, N.~D.
\newblock Gaussian processes for big data.
\newblock In \emph{Uncertainty in Artificial Intelligence}, pp.\  282, 2013.

\bibitem[Hern{\'a}ndez-Lobato \& Adams(2015)Hern{\'a}ndez-Lobato and
  Adams]{hernandez2015probabilistic}
Hern{\'a}ndez-Lobato, J.~M. and Adams, R.
\newblock Probabilistic backpropagation for scalable learning of {B}ayesian
  neural networks.
\newblock In \emph{International Conference on Machine Learning}, pp.\
  1861--1869, 2015.

\bibitem[Hern{\'a}ndez-Lobato et~al.(2014)Hern{\'a}ndez-Lobato, Hoffman, and
  Ghahramani]{hernandez2014predictive}
Hern{\'a}ndez-Lobato, J.~M., Hoffman, M.~W., and Ghahramani, Z.
\newblock Predictive entropy search for efficient global optimization of
  black-box functions.
\newblock In \emph{Advances in Neural Information Processing Systems}, pp.\
  918--926, 2014.

\bibitem[Houlsby et~al.(2011)Houlsby, Husz{\'a}r, Ghahramani, and
  Lengyel]{houlsby2011bayesian}
Houlsby, N., Husz{\'a}r, F., Ghahramani, Z., and Lengyel, M.
\newblock Bayesian active learning for classification and preference learning.
\newblock \emph{arXiv preprint arXiv:1112.5745}, 2011.

\bibitem[Kingma \& Welling(2013)Kingma and Welling]{kingma2013auto}
Kingma, D.~P. and Welling, M.
\newblock Auto-encoding variational {B}ayes.
\newblock \emph{arXiv preprint arXiv:1312.6114}, 2013.

\bibitem[Kirsch et~al.(2019)Kirsch, van Amersfoort, and
  Gal]{kirsch2019batchbald}
Kirsch, A., van Amersfoort, J., and Gal, Y.
\newblock Batchbald: Efficient and diverse batch acquisition for deep bayesian
  active learning, 2019.

\bibitem[Kuleshov et~al.(2018)Kuleshov, Fenner, and
  Ermon]{kuleshov2018accurate}
Kuleshov, V., Fenner, N., and Ermon, S.
\newblock Accurate uncertainties for deep learning using calibrated regression.
\newblock In \emph{International Conference on Machine Learning}, pp.\
  2796--2804, 2018.

\bibitem[Kulis et~al.(2006)Kulis, Sustik, and Dhillon]{kulis2006learning}
Kulis, B., Sustik, M., and Dhillon, I.
\newblock Learning low-rank kernel matrices.
\newblock In \emph{Proceedings of the 23rd international conference on Machine
  learning}, pp.\  505--512, 2006.

\bibitem[Lakshminarayanan et~al.(2017)Lakshminarayanan, Pritzel, and
  Blundell]{lakshminarayanan2017simple}
Lakshminarayanan, B., Pritzel, A., and Blundell, C.
\newblock Simple and scalable predictive uncertainty estimation using deep
  ensembles.
\newblock In \emph{Advances in Neural Information Processing Systems}, pp.\
  6402--6413, 2017.

\bibitem[MacKay(1992)]{mackay1992information}
MacKay, D.~J.
\newblock Information-based objective functions for active data selection.
\newblock \emph{Neural computation}, 4\penalty0 (4):\penalty0 590--604, 1992.

\bibitem[Martens \& Grosse(2015)Martens and Grosse]{martens2015optimizing}
Martens, J. and Grosse, R.
\newblock Optimizing neural networks with kronecker-factored approximate
  curvature.
\newblock In \emph{International conference on machine learning}, pp.\
  2408--2417, 2015.

\bibitem[Neal(1995)]{neal1995bayesian}
Neal, R.~M.
\newblock \emph{Bayesian Learning for Neural Networks}.
\newblock PhD thesis, University of Toronto, 1995.

\bibitem[Neal et~al.(2011)]{neal2011mcmc}
Neal, R.~M. et~al.
\newblock {MCMC} using {H}amiltonian dynamics.
\newblock \emph{Handbook of Markov Chain Monte Carlo}, 2\penalty0 (11), 2011.

\bibitem[Rahimi \& Recht(2008)Rahimi and Recht]{rahimi2008random}
Rahimi, A. and Recht, B.
\newblock Random features for large-scale kernel machines.
\newblock In \emph{Advances in Neural Information Processing Systems}, pp.\
  1177--1184, 2008.

\bibitem[Riquelme et~al.(2018)Riquelme, Tucker, and Snoek]{riquelme2018deep}
Riquelme, C., Tucker, G., and Snoek, J.
\newblock Deep {B}ayesian bandits showdown: An empirical comparison of
  {B}ayesian deep networks for thompson sampling.
\newblock In \emph{International Conference on Learning Representations}, 2018.

\bibitem[Shi et~al.(2019)Shi, Khan, and Zhu]{shi2019scalable}
Shi, J., Khan, M.~E., and Zhu, J.
\newblock Scalable training of inference networks for gaussian-process models.
\newblock \emph{arXiv preprint arXiv:1905.10969}, 2019.

\bibitem[Snoek et~al.(2019)Snoek, Ovadia, Fertig, Lakshminarayanan, Nowozin,
  Sculley, Dillon, Ren, and Nado]{snoek2019can}
Snoek, J., Ovadia, Y., Fertig, E., Lakshminarayanan, B., Nowozin, S., Sculley,
  D., Dillon, J., Ren, J., and Nado, Z.
\newblock Can you trust your model's uncertainty? evaluating predictive
  uncertainty under dataset shift.
\newblock In \emph{Advances in Neural Information Processing Systems}, pp.\
  13969--13980, 2019.

\bibitem[Srivastava et~al.(2014)Srivastava, Hinton, Krizhevsky, Sutskever, and
  Salakhutdinov]{srivastava2014dropout}
Srivastava, N., Hinton, G., Krizhevsky, A., Sutskever, I., and Salakhutdinov,
  R.
\newblock Dropout: a simple way to prevent neural networks from overfitting.
\newblock \emph{The journal of machine learning research}, 15\penalty0
  (1):\penalty0 1929--1958, 2014.

\bibitem[Sun et~al.(2018)Sun, Zhang, Wang, Zeng, Li, and
  Grosse]{sun2018differentiable}
Sun, S., Zhang, G., Wang, C., Zeng, W., Li, J., and Grosse, R.
\newblock Differentiable compositional kernel learning for {G}aussian
  processes.
\newblock In \emph{International Conference on Machine Learning}, pp.\
  4828--4837, 2018.

\bibitem[Sun et~al.(2019)Sun, Zhang, Shi, and Grosse]{sun2019functional}
Sun, S., Zhang, G., Shi, J., and Grosse, R.
\newblock Functional variational bayesian neural networks.
\newblock \emph{arXiv preprint arXiv:1903.05779}, 2019.

\bibitem[Swersky et~al.(2013)Swersky, Snoek, and Adams]{swersky2013multi}
Swersky, K., Snoek, J., and Adams, R.~P.
\newblock Multi-task bayesian optimization.
\newblock In \emph{Advances in neural information processing systems}, pp.\
  2004--2012, 2013.

\bibitem[Titsias(2009)]{titsias2009variational}
Titsias, M.
\newblock Variational learning of inducing variables in sparse {Gaussian}
  processes.
\newblock In \emph{Artificial Intelligence and Statistics}, pp.\  567--574,
  2009.

\bibitem[Wang \& Jegelka(2017)Wang and Jegelka]{wang2017max}
Wang, Z. and Jegelka, S.
\newblock Max-value entropy search for efficient {B}ayesian optimization.
\newblock In \emph{International Conference on Machine Learning}, pp.\
  3627--3635, 2017.

\bibitem[Yu et~al.(2006)Yu, Bi, and Tresp]{yu2006active}
Yu, K., Bi, J., and Tresp, V.
\newblock Active learning via transductive experimental design.
\newblock In \emph{Proceedings of the 23rd international conference on Machine
  learning}, pp.\  1081--1088. ACM, 2006.

\bibitem[Zhang et~al.(2018)Zhang, Sun, Duvenaud, and Grosse]{zhang2017noisy}
Zhang, G., Sun, S., Duvenaud, D., and Grosse, R.
\newblock Noisy natural gradient as variational inference.
\newblock In \emph{International Conference on Machine Learning}, pp.\
  5852--5861, 2018.

\end{thebibliography}
